\documentclass[lettersize,journal]{IEEEtran}
\usepackage{amsmath,amsfonts}
\usepackage{algorithmicx}
\usepackage{algorithm}  
\usepackage{algpseudocode}  
\usepackage{amsmath}  
\usepackage{array}
\usepackage[caption=false,font=normalsize,labelfont=sf,textfont=sf]{subfig}
\usepackage{textcomp}
\usepackage{booktabs}
\usepackage{stfloats}
\usepackage{url}
\usepackage{verbatim}
\usepackage{graphicx}
\usepackage{bbding}
\def\BibTeX{{\rm B\kern-.05em{\sc i\kern-.025em b}\kern-.08em
    T\kern-.1667em\lower.7ex\hbox{E}\kern-.125emX}}
\usepackage{balance}
\begin{document}
\title{Efficient Mixture-of-Expert for Video-based Driver State and Physiological Multi-task Estimation in Conditional Autonomous Driving}

\author{Jiyao Wang,~\IEEEmembership{Student Member,~IEEE,} Xiao Yang, Zhenyu Wang, Ximeng Wei, Ange Wang, Dengbo He, Kaishun Wu~\IEEEmembership{Fellow,~IEEE}
\thanks{Manuscript created October 2025; This work was supported by the National Natural Science Foundation of China (No. 52202425), Guangzhou Municipal Science and Technology Project (No. 2023A03J0011), and Guangzhou Science and Technology Program City-University Joint Funding Project (No. 2023A03J0001) (Corresponding author: Dengbo He).

Jiyao Wang, Xiao Yang, Zhenyu Wang, Ange Wang, and Dengbo He is with the Systems Hub, the Hong Kong University of Science and Technology (Guangzhou), Guangzhou, China,
(e-mail: jwanggo@connect.ust.hk; xyang856@connect.hkust-gz.edu.cn; zwang209@connect.hkust-gz.edu.cn; awang324@connect.hkustgz.edu.cn; dengbohe@hkust-gz.edu.cn). Ximeng Wei is with the University of Hong Kong, Hong Kong SAR, China,
(e-mail: aasimon0827@gmail.com). Kaishun Wu is with the Information Hub, the Hong Kong University of Science and Technology (Guangzhou), Guangzhou, China, (e-mail: wuks@hkust-gz.edu.cn).}}

\markboth{Journal of \LaTeX\ Class Files,~Vol.~18, No.~9, September~2020}%
{How to Use the IEEEtran \LaTeX \ Templates}

\maketitle

\begin{abstract}
Road safety remains a critical challenge worldwide, with approximately 1.19 million fatalities annually attributed to traffic accidents, often due to human error. As we advance towards higher levels of vehicle automation, challenges still exist, as driving with automation can cognitively over-demand drivers if engaging in non-driving-related tasks (NDRTs), or lead to drowsiness if driving is the sole task. This calls for the urgent need for an effective Driver Monitoring System (DMS) that can evaluate cognitive load and drowsiness in SAE Level-2/3 autonomous driving contexts. In this study, we propose a novel multi-task DMS, termed VDMoE, which leverages RGB video input to monitor driver states non-invasively. By utilizing key facial features to minimize computational load and integrating remote Photoplethysmography (rPPG) for physiological insights, our approach enhances detection accuracy while maintaining efficiency. Additionally, we optimize the Mixture-of-Experts (MoE) framework to accommodate multi-modal inputs and improve performance across different tasks. A novel prior-inclusive regularization method is introduced to align model outputs with statistical priors, thus accelerating convergence and mitigating overfitting risks. We validated our method by creating a new dataset (MCDD), which comprises RGB video and physiological indicators from 42 participants, as well as two public datasets. Our findings demonstrate the effectiveness of VDMoE in monitoring driver states, contributing to safer autonomous driving systems. Code and data are in https://github.com/WJULYW/VDMoE.
\end{abstract}


\section{Introduction}
\IEEEPARstart{R}{oad} safety is still an extreme challenge for societies \cite{kaplan2015driver,peng2018vehicle}. The World Health Organization (WHO) reported that approximately 1.19 million people die each year as a result of road traffic accidents, with human error being a significant contributing factor \cite{who2023global}. While autonomous technologies promise to reduce human error—the leading cause of traffic accidents—there remains a critical period where human drivers must share control with automated systems. Particularly, at the Society of Automotive Engineers (SAE) Level-2 (L2) automation \cite{sae2018taxonomy}, the vehicle can control both steering and acceleration/deceleration but still requires the driver to remain actively engaged and ready to take over at any moment; while with SAE Level-3 (L3) automation, the vehicle can handle most driving tasks but still requires human drivers to step in when prompted by the take-over requests. These levels of automation shift the driver's role away from continuous manual operation. Specifically, with L2 automation, the driver must remain actively engaged in monitoring the driving environment, even if their hands may be off the steering wheel. With L3 automation, the driver can disengage from driving tasks but must be ready to intervene when requested. Thus, drivers may experience low cognitive load when driving is their sole task, or excessive cognitive load when they are also engaged in non-driving-related tasks (NDRTs); both states can harm driving performance \cite{baumler1994validity, perello2021driver}. Cognitive overload from multi-dimensional resource demands can narrow the field of vision \cite{recarte2000effects} and reduce the ability to anticipate potential hazards \cite{muhrer2011effect}. In addition, long-term driving drowsiness and high cognitive load due to NDRTs may co-occur, further impairing driving performance \cite{soares2020analyzing, he2019high}. Therefore, a Driver Monitoring System (DMS) that can estimate these two states is critical for improving the safety and reliability of L2/L3 vehicles.

DMS has been widely studied in recent years. Traditional DMS relied on various sensors, including physiological sensors (e.g., Electrocardiography (ECG) sensors on the steering wheel) and vehicle-based sensors (e.g., steering angle, lane departure warnings) \cite{doudou2020driver}. However, in SAE L2 or L3 vehicles, the driver's manual driving performance data is not available most of the time, as the vehicle is controlled by the driving automation. Physiological signals, though can be served as effective indicators of drivers' state \cite{meteier2021classification, qu2023driver}, usually depend on invasive physiological sensors (e.g., electroencephalography (EEG) \cite{hu2024eeg, he2022classification}, electrooculography (EOG), \cite{picot2011line}) and thus are currently not practically applicable in vehicle cabins. Although there were some attempts to leverage various non-invasive physiological sensors \cite{li2015smartwatch, murugan2020detection}, drivers were still required to wear inconvenient
and costly signal acquisition equipment, which is still far from large-scale commercialization. Thus, considering the feasibility and cost of real-world in-vehicle deployment, camera is still a compelling alternative to monitor driver states non-intrusively \cite{yang2024video}. Especially, the fast development in deep computer vision enables efficient non-contact state detection, and thus, in recent years, several facial image-based driver drowsiness detection methods have been proposed \cite{sikander2020novel, mou2021isotropic}. 

However, most previous methods rely on single-frame detection, which ignores temporal variations in both surface facial expressions and internal physiological signals. In contrast, video-based approaches \cite{peng2024multi, yang2024video, huang2024self} can exploit continuous temporal features and potentially improve detection accuracy. While they still have several limitations: (1) some video-based drowsiness detection methods \cite{huynh2017detection} directly use continuous frames as input, leading to high computational cost and interference from redundant information; (2) some methods \cite{yang2024video, du2020vision} replace raw video with selected facial features (e.g., landmarks, eye and mouth regions), but cannot assess physiological states without additional sensor signals; and (3) current video-based DMSs mainly address single-task estimation of either drowsiness \cite{peng2024multi, yang2024video, huang2024self} or distraction \cite{hasan2024vision}, even though driver states interact and rarely occur in isolation \cite{wang2024revisiting}. For example, in conditional autonomous driving (SAE L2/L3), cognitive load can affect both drowsiness \cite{ayas2023drowsiness} and physiological responses \cite{wang2024association}. Yet, few non-contact solutions jointly estimate cognitive load, drowsiness, and physiological responses. Because training separate models for each task increases deployment costs and reduces iterative efficiency \cite{wang2024physmle}, there is an urgent need for a unified model that fully leverages the associations among multiple driver states.

\begin{table*}[t]
\setlength{\tabcolsep}{1mm}
\centering
\scriptsize
\caption{Comparison of Public Non-synthetic Dataset for State Estimation.}
\begin{tabular}{cccccccccc}
\toprule 
\textbf{Dataset} & \textbf{Date} & \textbf{Participant} & \textbf{Camera Setting} & \textbf{HR} & \textbf{RR} & \textbf{Drowsiness} & \textbf{cognitive load} & \textbf{Application Scenario} \\ 
\midrule
DROZY \cite{massoz2016ulg} & 2016 & 14 & NIR & \Checkmark & \XSolid & \Checkmark & \XSolid & - \\ 
NTHU \cite{weng2017driver} & 2016 & 36 & RGB, IR & \XSolid & \XSolid & \Checkmark & \XSolid & Driving \\ 
UTA-RLDD \cite{ghoddoosian2019realistic} & 2019 & 60 & RGB & \XSolid & \XSolid & \Checkmark & \XSolid & - \\ 
DMD \cite{jOrtega2020} & 2020 & 37 & RGB, IR, Depth & \XSolid & \XSolid & \Checkmark & \XSolid & Driving \\ 
FatigueView \cite{yang2022FatigueView} & 2022 & 95 & RGB, IR & \XSolid & \XSolid & \Checkmark & \XSolid & Driving \\ 
MOCAS \cite{jo2024mocas} & 2024 & 21 & RGB & \Checkmark & \XSolid & \XSolid & \Checkmark & - \\ 
\midrule 
MCDD & 2024 & 42 & RGB & \Checkmark & \Checkmark & \Checkmark & \Checkmark & Driving \\ 
\bottomrule 
\end{tabular}
\textnormal{\\Notes: NIR means the Near Infrared camera, IR means the Infrared camera.}
\vspace{-5mm}
\label{t0}
\end{table*}

To address the aforementioned issues, we propose an RGB video-based multi-task DMS (VDMoE). Specifically, we assume the driver's face is largely visible to the camera. Then, based on previous studies \cite{yang2024video, yang2023real}, we first utilized the key facial features to reduce redundant information and computational costs. At the same time, considering cardiac and respiration activities are crucial predictors of driver states \cite{wang2024association, ayas2023drowsiness}, and are also important driver health indicators, we integrated the remote Photoplethysmography (rPPG) technology \cite{verkruysse2008remote} into our model. Since the facial landmark and local eye and mouth region processes contain limited temporal color change of the facial skin pixel points, which can be converted to a change of the blood volume under the skin \cite{wang2024physmle}, we first selected key facial regions of interest based on facial landmarks. Then, by transforming the color space (from RGB to YUV) as well as by band-pass denoising, we obtained the alternative multi-modal information input Spatial-temporal Map (STMap). Next, to fully leverage the dependencies between different tasks, the Mixture-of-Experts (MoE) \cite{shazeer2016outrageously} structure is introduced. Being different from the classic MoE structure, the heterogeneous gating mechanism and spatio-temporal expert separation are designed for the multi-modal alternative inputs to improve the multi-task performance. Besides, recent video-based DMS used various types of neural networks (e.g., CNN \cite{peng2024multi}, Transformer \cite{yang2024video}) that are suitable for processing different types of data or extracting different levels of feature information \cite{hunter2012selection}. However, given the demand for real-time assessment of DMS, networks with excessive parameters and computational complexity will pose a challenge to deployment hardware. Therefore, we instantiated the nonlinear feature learning component of the basic block of VDMoE as a simple two-layer multi-layered perception (MLP) network to replace the previous modules represented by residual convolution \cite{he2016deep} or multi-head attention \cite{vaswani2017attention}. Lastly, for the optimization goal, to explicitly capture the dependency between different states, we strengthen the learning capacity of VDMoE through one regularization based on prior knowledge from human factors fields \cite{ayas2023drowsiness}. Given there is no proper dataset for multi-task video-based DMS development that considers both the cognitive load and drowsiness in the driving automation context (see Table \ref{t0}), a driving simulator experiment with 42 participants was conducted. Extensive experiments based on our dataset illustrate the effectiveness of our method. In all, the contributions of our work are summarized as follows:
\begin{itemize}
    \item A multi-task RGB video-based driver state monitoring method (VDMoE) is proposed in this work. This system concentrates on the drowsiness and physiological reflection moderated by the multi-dimensional cognitive load in the context of driving automation. As far as we know, this is the first multi-task model that can simultaneously estimate driver drowsiness, cognitive load, and cardiac and respiratory activities.
    \item To achieve the balance between estimation accuracy and efficiency, we replaced the full video with key facial features (landmarks, eye, and mouth area) as input, and introduced the STMap representing periodical facial color changes to supplement alternative physiological features.
    \item Inspired by the superior performance of MoE in multi-task learning, we further optimized the MoE structure to accommodate the multi-modal and spatio-temporal input through a heterogeneous gating mechanism and spatio-temporal expert separation. At the same time, we replaced more complex network structures with MLPs for lightweight and future deployment. Comparison of computational cost with existing baselines proves our proposals' effectiveness.
    \item  A novel prior-inclusive regularization, inspired by human factor engineering studies, was proposed to enforce the multi-task model's outputted probability distributions for each state to be aligned with the statistical prior distributions so that we can speed up convergence and reduce the risk of overfitting the model to particular individuals. Results from the ablation test validate its effectiveness.
    \item To validate the effectiveness of our method, the first multi-modal cognitive load and drowsiness driving dataset (MCDD) in the L3 context was created. In total, 42 participants were involved in this dataset, which included about 105,840 seconds RGB video and cardiac and respiration-related physiological indicators. The cleaned processed dataset will be released.
\end{itemize}

\section{Related Works}
\subsection{Physiological Signal-Based Driver State Monitoring}
Driver drowsiness \cite{saxby2013active} can negatively affect drivers' performance when driving with automation, especially in takeover events. Thus, driver state monitoring is necessary to ensure safe driving in the context of driving automation \cite{qu2023driver}. According to \cite{wang2025towards,deng2024analysis}, in conditional autonomous vehicles, certain physiological metrics can be strongly correlated with driver state. Therefore, researchers proposed various physiological signal-based driver state monitoring methods. Commonly used psychophysiological measures include electrophysiological signals and human physical movements \cite{lohani2019review,wang2024cognitive}. For example, Cui et al. \cite{cui2022eeg} designed an interpretable convolutional neural network for drowsiness detection using EEG signals. In addition, due to the difficulty of obtaining EGG signals, methods based on ECG and Electrodermal Activity (EDA) signals \cite{wang2026drowsydg,cella2010measuring} have also been widely proposed. Yang et al. \cite{yang2023real} developed an attention-enabled recognition network with a decision-level fusion architecture to estimate cognitive load based on physiological data. Another study used physiological signals and facial images to detect driver drowsiness \cite{peng2024multi}. However, previous studies relied on intrusive sensors that had to contact the driver's body, making them impractical for real driving environments. To address this, the VDMoE proposed in this paper uses a non-contact RGB camera to estimate multiple driver states, replacing traditional contact-based monitoring.

\subsection{Remote Physiological Measurement}
Since 2008, Verkruysse et al. \cite{verkruysse2008remote} proposed the rPPG technique for detecting physiological data using only a single consumer-grade camera. Nowadays, the rPPG technique has become a mainstream method in remote physiological data monitoring because of its non-contact characteristics. For example, to improve the robustness of pulse rate measurements, Hann et al. \cite{de2013robust} proposed an analytical approach to tackle the motion problem in rPPG measurement. Wang et al. \cite{wang2016algorithmic} introduced a mathematical model to increase understanding of the rPPG and designed a new algorithm to measure the heart rate. Tarassenko et al. \cite{tarassenko2014non} used autoregressive modeling to implement respiratory rate monitoring. However, traditional rPPG methods require manual adjustment of inaccurate data and are mostly limited to a single task. 

Recently, the combination of deep learning algorithms and rPPG technology has effectively improved accuracy in complex environments \cite{lu2021dual,yu2023physformer++}. Many studies have been conducted in the direction of multi-task monitoring of physiological data. For example, Liu et al. \cite{liu2020multi} presented a multi-task network to enable respiration and heart rate measurements on a mobile platform. Narayanswamy \cite{narayanswamy2024bigsmall} et al. proposed a model based on facial movement, HR, and respiration measurement. Wang et al. \cite{wang2024physmle} designed a PhysMLE model to simultaneously calculate HR, heart rate variability, RR, and blood oxygen saturation. Nevertheless, existing multi-task rPPG approaches typically focus on physiological signal estimation (e.g., HR and RR) or single psychological indicators \cite{wang2025align}, rather than jointly inferring both physiological and cognitive/drowsiness states. Consequently, they lack the robustness and comprehensive capability for real-time monitoring under Level-2/3 automation, which our VDMoE model explicitly addresses.

\subsection{Non-contact Driver State Monitoring}
Over the past few decades, advances in deep learning and growing demands for driving safety have made non-contact driver state monitoring mainstream. Vision-based non-contact methods are typically divided into conventional criterion-based approaches and direct deep learning–based approaches. Criterion-based methods use handcrafted indicators—such as eye opening degree, mouth opening degree, or head pose—combined with traditional algorithms to detect driver states \cite{bai2021two}, but they generalize poorly to diverse, complex driving scenarios. As a result, direct deep learning–based drowsiness detection has become more prevalent. Ma et al. \cite{su2018yawning} proposed an image-based yawn detection algorithm that takes a single driver image as input, but it cannot exploit the temporal information in continuous video. To address this, several video-based deep learning models have been proposed \cite{yang2020driver,park2016driver,zhao2020driver}, though they incur higher training and deployment costs. More recently, Mou et al. \cite{mou2021isotropic} introduced a self-supervised algorithm that uses eye, nose, and head optical flow for drowsiness detection. Other studies \cite{bai2021two,cheng2019assessment} rely on facial landmarks. For example, Yang et al. \cite{yang2024video} used eye, mouth, and other facial feature points as inputs to estimate driver drowsiness levels.

While recent video-based DMS and rPPG research have begun exploring multi-task architectures, particularly those based on deep CNNs or Transformer backbones, these approaches still face substantial parameter overhead as tasks are added. Existing multi-task rPPG methods often suffer from model bloat due to task-specific decoupling layers and conflicting optimization objectives, leading to a `seesaw effect' between tasks \cite{wang2024physmle,narayanswamy2024bigsmall}. Moreover, to our knowledge, no non-contact video-based algorithm has concurrently addressed cognitive load assessment, drowsiness detection, and physiological monitoring in Level-2/3 driving. In contrast, VDMoE employs a MoE-style architecture well suited to multi-task learning: a small set of shared expert MLPs is dynamically routed by an element-wise gating network, and spatio-temporal features are handled by specialized experts. This design lets new tasks be added without a proportional increase in parameters. Using lightweight two-layer MLP experts instead of heavy CNN or Transformer modules \cite{yu2023physformer++,liu2020multi}, VDMoE achieves a smaller model size, lower inference latency, and integrates rPPG to capture physiological information. VDMoE thus outperforms existing Transformer- and CNN-based multi-task models in efficiency and accuracy, while uniquely providing a non-contact solution for simultaneous estimation of driver drowsiness, cognitive load, and physiological states in L2/3 automation.

\begin{table}[h]
\caption{A summary of symbols and descriptions}\label{t1}
\begin{tabular}{c|l} 
Symbol & Description\\
\hline
$B$ & Number of samples in one batch.\\
$T$ &Number of frames of each video.\\
$N_f$ & Number of facial landmark points in each frame.\\
$x_i,y_i$ & The horizontal and vertical coordinates of a facial landmark.\\
$L, W, C$ & Length, width, and the number of channels of facial regions.\\
$L_e, W_e$ & Length, width of facial regions of eyes.\\
$L_m, W_m$ & Length, width of facial regions of mouths.\\
$I_l, I_r, I_m$ & The local area of left eyes, right eyes, and mouths\\
$N_s$ & Number of ROIs of STMaps.\\
$I_l$ & The local facial regions of drivers.\\
$I_s$ & STmaps.\\
$I_f$ & The facial landmarks matrix.\\ 
$X$ & Model inputs.\\
$f(*)$ & The target multi-task model.\\
$\theta$ & Learnable parameters in the model.\\
$P(*)$ & The task-specific estimation heads.\\
$h_i$ & The ground-truth HR .\\
$r_i$ & The ground-truth RR.\\
$S$ & Size of the sliding step.\\
$L$ & Number of VDMoE block.\\
$K$ & Number of experts in the VDMoE block.\\
$D$ & Spatial dimension of each feature embedding.\\
$V$ & Spatial dimension of the combined feature embedding.\\
$m_t$ & Feature vector maintains the temporal information.\\
$m_s$ & Feature vector with spatial information.\\
$E(*)$ & The expert network. \\
$R(*)$ & The expert router network. \\
$G(*)$ & The task-related feature gating network.\\
$\tau$ & Temperature factor for soft regularization.\\
$\lambda$ & Adaptation factor for regularization term.\\
$k$ & Trade-off parameter.\\
\\
\end{tabular}\\
\vspace{-5mm}
\end{table}
 
\section{Methodology}
\subsection{Problem Formulation}

From previous studies in human factors \cite{ayas2023drowsiness, wang2024association}, we know that both drivers' drowsiness and cognitive load are not only highly correlated with each other, but are also significantly associated with facial features (e.g., the degree of opening and closing of the eyes and mouth) and physiological features (e.g., HR, RR). Therefore, integrating state and physiological estimation into one multi-task model can effectively exploit correlations between tasks and features while compressing the computational cost further.

Inspired by this, we introduced a multi-task learning architecture, a video-based multi-task DMS mixture of experts model (VDMoE). As shown in Figure \ref{f3}, suppose we have a batch of $B$ raw videos. Being different from previous methods taking 3D videos as input \cite{yang2020driver, mou2021isotropic}, we extracted heterogeneous facial features from videos as input $X$. Firstly, following \cite{yang2024video}, we extracted facial landmarks, and subregion facial videos (i.e., eyes and mouth) from the video. Then, as only movement features from landmarks and above subregions might be insufficient for physiological measurement, we combined the knowledge from rPPG \cite{wang2023hierarchical, wang2024diff} to supplement skin light changes into the input $X$. The specific preprocessing steps are in the following section. Finally, given $X$, we tried to obtain both physiological signals $\{h_i,r_i\}_{i=1}^{N}$ and state estimation $\{d_i,c_i\}_{i=1}^{N}$ by training a multi-task model $f(X;\theta)$, where $h_i$ is the heart rate, $r_i$ is the respiratory rate, $d_i$ is drowsiness levels and $c_i$ is the cognitive load levels. The trainable parameter $\theta$ is our optimization objective. Finally, the multi-task estimation target is $Y = \{h, r, d, c\}$

\begin{figure*}
\begin{center}
\includegraphics[scale=0.35]{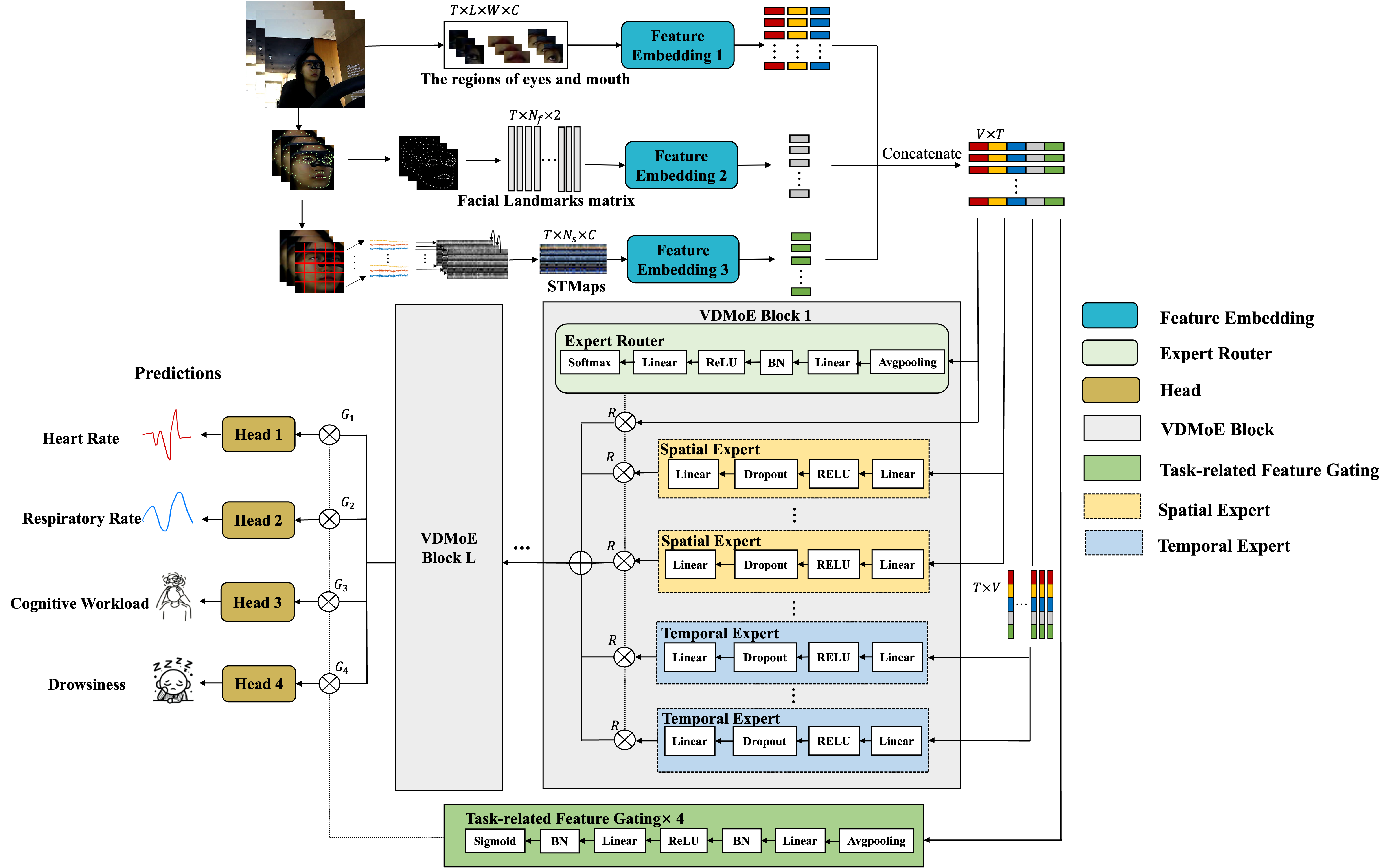}\\
\end{center}
\vspace{-4mm}
\caption{This figure presents the proposed multi-task driver state monitoring method VDMoE. We first transform videos into STMaps, facial landmarks matrix, and facial regions. Then inputs them into the model to output physiological signals, drowsiness, and cognitive load.}\label{f1}
\vspace{-4mm}
\end{figure*}

\subsection{Preprocess}

In the original dataset, each video contains $T$ frames. To ensure a consistent input size, we apply a sliding window to segment and augment the raw dataset, generating additional samples with a fixed frame count and reducing overfitting. All video samples are set to $T$ frames and obtained from the original videos using a sliding window with step size $S$.

At the beginning of preprocessing, we first applied the facial detection methods to each frame of the input video to generate 2D coordinates of $N_f$ facial landmarks with $N_f$ key points. Since the length of each video is $T$ frames and each facial landmark is represented by horizontal and vertical coordinates, denoted as $(x_i,y_i)$, each video can be represented as a facial landmarks matrix of shape $T\times N_f\times 2$. Then, by using the outer boundaries corresponding to the facial landmarks as bounding boxes, the regions of the left eye, right eye, and mouth $I_l, I_r, I_m$  were obtained from each original facial video frame. To standardize the input, the subregion videos of the left eye and right eye $I_l, I_r$ are resized to $T\times L_e\times W_e\times C$, where $L_e, W_e$ are the length, width of the subregion videos corresponding to eyes. Similarly, the subregion video of mouth $I_m$ is the size of $T\times L_m\times W_m\times C$, where $L_m, W_m$ are the length, and width of the mouth video. Subsequently, the entire facial areas were cropped into $N_s$ regions of interest (ROIs) \cite{yang2025consistencyenhancetesttimeadaptation}. These ROIs were converted from RGB to YUV. Then, since we know that the skin light changes owing to the volume and oxygen saturation of the blood correspond to the low amplitude of the low-frequency component \cite{niu2018automatic}, we used the first-order difference to remove the large amplitude caused by head movements and changes in light. After a Butterworth filter with parameter [0.4,10] Hz to retain the frequency band of interest \cite{wang2025physdrive}, we processed the ROIs into a spatial-temporal map (STMap) $I_s \in \mathbb{R}^{T\times N_s\times C}$ to measure physiological signals. Among them, the channel number $C=3$ is constant. In all, the model's input $X$ can be described as $X = \{I_l, I_r, I_m, I_f, I_s\}$.

\subsection{VDMoE}

\subsubsection{Feature Embedding Modules}

Due to the different shapes of the heterogeneous facial features (i.e., STMap $I_s$, facial landmarks matrix $I_s$, and the regions of the left eye $I_l$, right eye $I_r$, and mouth $I_m$), we designed feature extraction modules tailored to ensure that facial features from different types of inputs are encoded to the same size of high dimension space. The specific architecture and parameters of the network are listed in Table \ref{t_emb}. 

Specifically, we built separate convolution and pooling pipelines for each frame of the facial subregion videos, enabling parallel feature processing across frames while preserving temporal information. We first applied a dimensionality-increasing operation, then a convolution to maintain the matrix’s temporal order. For STMap feature embedding, because STMaps encode both temporal and spatial information, we designed convolutions that preserve their spatial dimensions and maintain correlations between STMap temporal features and other inputs. In the final embedding stage, we used adaptive pooling and reshaping to compress spatial dimensions into one-dimensional vectors, matching the output size of the other feature embeddings. Besides, we formed the facial landmark matrix by concatenating per-frame landmarks in chronological order, using a feature embedding strategy similar to that for the facial regions. Finally, a fully connected layer compressed all features into one-dimensional vectors with a fixed output shape.

Through feature embedding, each type of input facial feature was converted to a two-dimensional vector of shape $T\times D$. Because all transformations operate only on the spatial dimension (excluding the temporal dimension $T$), and $T$ is fixed during embedding, this step distills the spatial facial features of each frame into a vector of size $D$ while preserving temporal order. Concatenating these vectors along the last dimension yields the combined feature vector $m$ with shape $T \times 5D$, where we denote $5D$ as the spatial dimension $V$.  

\begin{table}[t]
\setlength{\tabcolsep}{1mm} 
\renewcommand{\arraystretch}{2} 
\centering
\scriptsize 
\caption{Network Structures of FeatureEmbedding of STMaps, Facial Landmarks Matrix and Facial Regions}
\begin{tabular}{cccc}
\toprule
\textbf{Layer}  & \textbf{Input size}                 & \textbf{Output size}                  & \textbf{Content}                                  \\ \midrule
\multicolumn{4}{c}{\textbf{Feature Embedding 1 of Facial Regions}}                                                                          \\ \midrule
Conv1        & \( B \times H \times W \times C \)             & \( B \times H \times W \times 16 \)            & Conv + ReLU                                       \\ 
Pool1        & \( B \times H \times W \times 16 \)            & \( B \times \dfrac{H}{2} \times \dfrac{W}{2} \times 16 \) & MaxPool                                           \\ 
Conv2        & \( B \times \dfrac{H}{2} \times \dfrac{W}{2} \times 16 \) & \( B \times \dfrac{H}{2} \times \dfrac{W}{2} \times 32 \) & Conv + ReLU                                       \\ 
Pool2        & \( B \times \dfrac{H}{2} \times \dfrac{W}{2} \times 32 \) & \( B \times \dfrac{H}{4} \times \dfrac{W}{4} \times 32 \) & MaxPool                                           \\ 

Reshape     & \( B \times \dfrac{H}{4} \times \dfrac{W}{4} \times 32 \)          & \( B \times (\dfrac{H}{4} \times \dfrac{W}{4} \times 32) \)          & Reshape                                          \\ 
FC           & \( B \times (\dfrac{H}{4} \times \dfrac{W}{4} \times 32) \)                             & \( B  \times 128 \)                        & Fully Connected                                 \\ \midrule
\multicolumn{4}{c}{\textbf{Feature Embedding 2 of STMaps}}                                                                          \\ \midrule
Conv1        & \( B \times N_s \times C \)           & \( B \times N_s \times 64 \)          & Conv + ReLU                                       \\ 
BatchNorm1   & \( B \times N_s \times 64 \)          & \( B \times N_s \times 64 \)          & BatchNorm                                         \\ 
Conv2        & \( B \times N_s \times 64 \)          & \( B \times N_s \times 128 \)         & Conv + ReLU                                       \\ 
AdaptivePool & \( B \times N_s \times 128 \)         & \( B \time 1 \times 128 \)         & AdaptivePool                                      \\ 
Reshape      & \( B  \time 1 \times 128 \)         & \( B \times 128 \)         & Reshape                                           \\ \midrule
\multicolumn{4}{c}{\textbf{Feature Embedding 3 of Facial Landmarks Matrix}}                                                                              \\ \midrule
Reshape1     & \( B \times N_f \times 2 \)           & \( B \times 1 \times N_f \times 2 \)           & Reshape                                          \\ 
Conv1        & \( B \times 1 \times N_f \times 2 \)           & \( B \times 32 \times N_f \times 1 \)          & Conv + ReLU                                     \\ 
BatchNorm1   & \( B \times 32 \times N_f \times 1 \)          & \( B \times 32 \times N_f \times 1 \)          & BatchNorm                                       \\ 
Reshape2     & \( B \times 32 \times N_f \times 1 \)          & \( B \times (32 \times N_f) \)          & Reshape                                          \\ 
FC1          & \( B \times (32 \times N_f) \)          & \( B \times 128 \)                    & Fully Connected                                  \\ \bottomrule
\end{tabular}
\label{t_emb}
\vspace{-5mm}
\end{table}

\subsubsection{VDMoE Block}
To learn the underlying correlations between different tasks, based on the classic MoE block \cite{shazeer2016outrageously}, we proposed the hierarchical VDMoE architecture to accommodate the spatio-temporal characteristics of heterogeneous facial features. There are $L$ VDMoE blocks in total. Specifically, as previous feature embedding modules have already encoded heterogeneous facial features to a high-dimensional space, to minimize the computational cost and avoid the decline in generalizability due to high complexity, we instantiated each expert in the VDMoE block with a two-layered MLP. Additionally, given our $m$ put the spatial information from heterogeneous facial features at the last dimension, following the matrix multiplication in linear algebra, the weights of MLP experts can only be applied to the last dimension (i.e., spatial information). However, the temporal information is also important \cite{bai2021two}, as both drivers' behavior and physiological reflection under specific states should not be instantaneous. Thus, we introduced the spatio-temporal expert mechanism in our VDMoE block. Specifically, given that only spatial information can be processed in $m$, we first transposed the $m$ at the last two dimensions and obtained the $m_t \in \mathbb{R}^{V\times T}$, where temporal information of each spatial channel can be learned. Then, in each VDMoE block, we constructed $K$ experts, which can be further equally classified as spatial experts $E_s(*)$ and temporal experts $E_t(*)$ corresponding to $m$ and $m_t$ respectively. Experts share no parameters and are initialized separately. The structure of each type of expert is shown in Figure \ref{f1}. Then, the $m$ was processed to $\frac{K}{2} \ m_{s}^{'}$ by $\frac{K}{2}$ different spatial experts $E_s(*)$. Similarly, we obtained $\frac{K}{2} \ m_{t}^{'}$ after input $m_t$ to $\frac{K}{2} E_t(*)$. Moreover, to avoid information degradation with the increase in the number of layers in neural networks \cite{he2016deep}, we leveraged the jump connection in the residual network. Therefore, the intermediate features $m^{'}$ consist of $\frac{K}{2}\  m_{s}^{'}$, $\frac{K}{2}\  m_{t}^{'}$, and the initial input of the VDMoE block $m$. The whole procedure is formalized as follows:

\begin{equation}
\begin{aligned}
m &\in \mathbb{R}^{T \times V}, \quad m_t = m^\top \in \mathbb{R}^{V \times T}, \\
m_{s}^{'} &= \{E_s^{i}(m)\}^{\frac{K}{2}},\ i\in[1,\frac{K}{2}],\\
m_{t}^{'} &= \{E_t^{i}(m_{t})\}^{\frac{K}{2}},\ i\in[\frac{K}{2}, K],\\
m^{'} &= \{m_{s}^{'},m_{t}^{'\top}, m \}.
\end{aligned}
\label{e1}
\end{equation}

Then, we designed an expert router mechanism $R(*)$ to dynamically aggregate the above intermediate features. Each router $R(*)$ corresponds to a specific intermediate feature. The router $R(*)$ took the $m$ as the input and nonlinearly converted it from $\mathbb{R}^{T\times V}$ to a probability vector of shape $(K+1)\times 1$. In detail, upon receiving the $m$, we first applied average pooling to compress the temporal dimension. Then, a two-layer MLP performed a nonlinear transformation, followed by a Softmax function to generate the output. Finally, the probability vector generated by the router was used to selectively activate features by weighting the intermediate variables. Each element of the vector represents the activation probability for a corresponding feature. By weighted summing these intermediate features with probability features, we emphasized important features while suppressing less relevant ones. The aggregated feature $m^{'}$ was sent to the next VDMoE block. 
\subsubsection{Task-Related Feature Gating and Estimation Head}

After the VDMoE blocks, a representative task-generic feature $m^{'} \in \mathbb{R}^{T\times V}$ was obtained. Although there are correlations among multiple target tasks and can be learned by one shared backbone, separate feature spaces for different tasks are still necessary \cite{wang2024physmle}. Therefore, we designed element-wise gated networks to select sparse high-level representations for specific tasks. Specifically, four element-wise gates $G_{i}(*)$ were initiated, which took the original embedding feature $m$ as input. Then, seeing Figure \ref{f1}, after the average pooling and a two-layered MLP, the output of each gate $G_{i}(*)$ was controlled by a Sigmoid function, which makes each element of the output vector in the range [0, 1]. We argue that, being different from Softmax-based weighted aggregation \cite{shazeer2016outrageously,riquelme2021scaling}, the element-wise Sigmoid-based gate can adaptively
control the importance of low-level information relative to specific tasks, thus selecting proper subspaces $\{m^{'}_i\}^{4} \in \mathbb{R}^{V}$ from a uniform task-generic space $m^{'}$. Lastly, we input these subspaces into four task-specific estimation heads as:

\begin{equation}
\begin{aligned}
Y &= P(G(m)\cdot Pool(m^{'})) \\
&=Linear(Sigmoid(MLP(Pool(m)))\cdot Pool(m^{'})).
\end{aligned}
\label{e2}
\end{equation}

\subsection{Prior-driven Alignment}

As noted by \cite{ayas2023drowsiness}, drowsiness and cognitive load tend to exhibit an overall negative correlation in driving contexts. This inverse relationship is rooted in cognitive-resources theory \cite{kahneman1973attention}, which indicates the limited human attentional capacity: low task demands can lead to mind-wandering and increased drowsiness \cite{hockey1997compensatory}. Conversely, moderate task demands optimally engage attentional control mechanisms, sustaining cortical arousal, maintaining vigilance, and delaying the onset of fatigue \cite{merlhiot2022drowsiness}. Studies of vigilance decrement further demonstrate that under-stimulation can produce lapses in attention and microsleeps \cite{warm2008vigilance}, whereas appropriately challenging tasks improve sustained performance. We therefore leverage this well-established theoretical and empirical foundation to introduce a prior-driven soft regularization $\mathcal{L}_{align}$, enforcing a general negative alignment between our drowsiness and cognitive workload estimates.

Specifically, given the estimated vector of drowsiness and cognitive load by corresponding heads, we applied the Softmax function to each vector to convert them to the binary probability vectors $p_{drow}$ and $p_{cog}$. Then, based on the Kullback-Leibler Divergence, we tried to maximize the distributional difference between $p_{drow}$ and $p_{cog}$ with Eq. (\ref{eq3}). Additionally, since this correlation may not hold in extreme cases (e.g., when cognitive load exceeds capacity and speeds up drowsiness \cite{ma2018relationship}), we introduce a temperature hyperparameter $\tau$ to scale the two probability distributions and control the sharpness of the output. We set $\tau$ as 2 to smooth the distributions:

\begin{equation}
\mathcal{L}_{align}=  -\sum_{B} \left(p_{drow}/\tau\right) \log\left(\frac{\left(p_{drow}/\tau\right)}{p_{cog}/\tau}\right)
\label{eq3}
\end{equation}

\subsection{Optimization Goal}
To facilitate the estimation of drowsiness, cognitive load, HR, and RR. We designed different optimization goals depending on the properties of each task. For the loss of drowsiness $\mathcal{L}_{drow}$ and cognitive load $\mathcal{L}_{cog}$, given there is strong subjective consciousness and individual differences in assessing drowsiness and cognitive load with the self-report questionnaires \cite{paulhus2007self}, participants will label their current status with varied standards. When training the model with data from different individuals, the mapping between facial features and state labels is not robust. Thus, we applied the generalizable cross-entropy loss (Truncated Loss) \cite{zhang2018generalized} to optimize the drowsiness and cognitive load estimation, and the smooth L1 loss for the estimation of HR $\mathcal{L}_{hr}$ and RR $\mathcal{L}_{rr}$ following \cite{wang2023hierarchical}. Besides, to suppress some meaningless effects of the regularization at the early iterations, we applied the adaptation factors $\lambda$ as Eq. (\ref{eq4})

\begin{equation}
\begin{aligned}
t = \frac{Iter_{current}}{Iter_{total}}, \lambda = \frac{2}{1+\exp{-10t}}.
\end{aligned}
\label{eq4}
\end{equation}

Then, we combined the multi-task optimization goals and the regularization term into one loss formula $\mathcal{L}_{overall}$ to conduct the joint training with one trade-off parameter $k_1$.

\begin{equation}
\mathcal{L}_{overall} = \mathcal{L}_{drow}+\mathcal{L}_{cog}+\mathcal{L}_{hr}+\mathcal{L}_{rr}+\lambda*k_1*\mathcal{L}_{align}.
\label{C5_eq5}
\end{equation}

\section{Materials}
\subsection{MCDD Dataset}
\begin{figure}
\begin{center}
\includegraphics[scale=0.55]{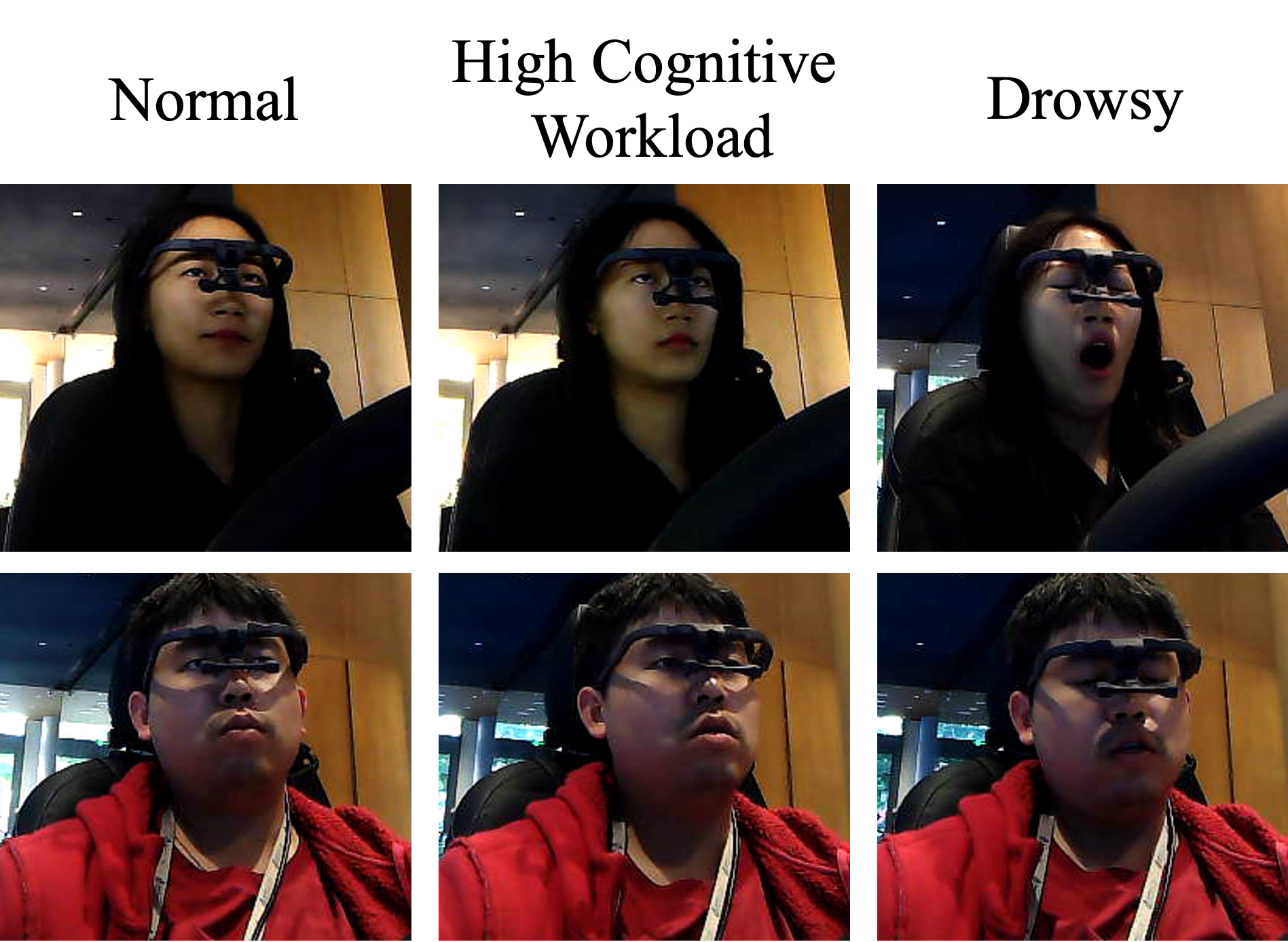}\\
\end{center}
\vspace{-4mm}
\caption{Sample RGB video frames of drivers in different states in MCDD (eye-tracking data is not used in this paper).}\label{f4}
\vspace{-4mm}
\end{figure}

To mitigate the gap in previous public datasets, we collected a new multi-modal cognitive load and drowsiness driving dataset (MCDD). Some example frames of drivers with different states and the distribution of our extracted HR and RR are shown in Figure \ref{f4}.

\subsubsection{Experiment design}
\begin{figure*}
\begin{center}
\includegraphics[scale=0.45]{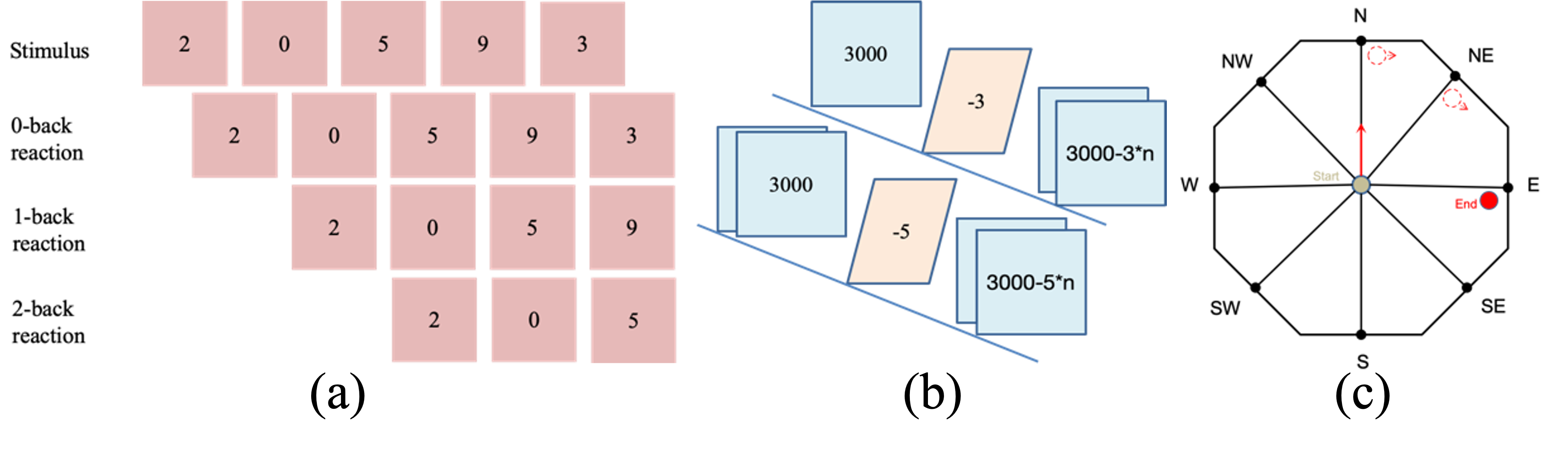}\\
\end{center}
\vspace{-8mm}
\caption{Cognitive task design, including (a) n-back task, (b)math task, and (c) spatial task.}\label{f3}
\vspace{-4mm}
\end{figure*}

Previous study \cite{liu1994mental} showed that drivers' cognitive resources are multi-dimensional, and different dimensions of non-driving-related tasks (NDRTs) can result in different levels of cognitive load. Given the moderating effect of drivers' cognitive load on drowsiness and physiological responses \cite{soares2020analyzing, wang2024association}, this driving simulator study used a within-subject design to investigate the impact of varying types of cognitive NDRT on drivers' states. As detailed in Table \ref{t1} and Figure \ref{f3}, three types of cognitive tasks, encompassing 6 specific tasks, were used in addition to a baseline condition without NDRTs, leading to 7 NDRT tasks. To systematically manipulate cognitive workload, we treated the driving-only condition (i.e., SAE L3 automated driving without any NDRT) as an underload baseline, representing the low cognitive demands of monotonous conditional automation. The N-back and math tasks then imposed incremental working memory and arithmetic demands: 0-back and counting backwards by 5 were medium-load conditions, whereas 2-back and counting backwards by 3 were high-load conditions due to their increased information-processing requirements \cite{wang2025impact}. The spatial task further imposed high visuo-spatial processing demands similar to those encountered when following complex navigation instructions \cite{wang2024revisiting}. 

Each participant completed three drives per NDRT task, resulting in a total of 21 drives (7 NDRTs * 3 drives). A Latin square design, with 21 unique orders, was implemented to mitigate potential order effects. All driving scenarios took place on simulated two-way, six-lane highways with a speed limit of 120 kilometers per hour and a traffic density of 6 vehicles per kilometer per lane. At the beginning of each drive, participants needed to first quickly manually drive to the middle lane. Then, the SAE L3 driving system was activated and set to keep a speed of 110 kilometers per hour. Participants were informed that they were allowed to take their hands off the steering wheel and finish NDRTs after activating the L3 driving automation. Each drive covered a distance of approximately 7 kilometers, corresponding to about 3.8–4.0 minutes of driving.

\begin{table*}[t]
\setlength{\tabcolsep}{1mm}
\centering
\caption{Summary of Cognitive Load Tasks.}
\begin{tabular}{clllc}
\toprule 
 \textbf{Task Type}    & \textbf{Description} & \textbf{Example}    & \textbf{Cognitive Load Level} & \textbf{Cognitive Resource}  \\
\midrule 
\textbf{N-back Task}  \cite{jaeggi2010concurrent}& A series of stimuli numbers are presented & 0-back (0-B), 1-back (1-B),       & Three levels; The difficulty     & Memory  \\
& with a pause between each. Participants & 2-back (2-B) tasks. &  increases with the increase&   \\
& recall and verbally report the stimulus&  &of n.&   \\
&  that is n positions earlier. &  & &   \\
\hline
\textbf{Math Task}  \cite{meteier2021classification}      & Oral backward counting from 3,000 by & Counting backward by 3 (MT1)  & Two levels; level of cognitive& Calculation   \\
& increments of 3 (non-integer) & or 5 (MT2) from 3,000.   & depends on the nature of the&  \\
& or 5 (integer). &   & math tasks.&\\
\hline
\textbf{Spatial Task} \cite{liang2010combining}   & Participants listen to an audio clip  & E.g., "What direction is this person        & One level: Simulates high & Spatial Processing\\
& describing a route and identify the & when he goes to the north station     & cognitive demands similar to  & \\
&main direction faced at the end.   & and moves two stations clockwise?"     &  those in navigation systems.    & \\
&  & (Answer: East) (ST)       &      & \\
\bottomrule 
\end{tabular}
\label{t1}
\end{table*}

Across all drives, this yields roughly 1.5 hours of continuous automated driving per participant within a 3-hour session. In line with previous simulator and L3 automated driving studies \cite{tong2024difference, figalova2023fatigue}, prolonged monotonous driving over 60–90 minutes reliably increases fatigue and sleepiness, especially under low workload and L3 automation conditions.  Participants were provided with pre-experiment instructions to ensure adequate rest and avoid substances that could impact their performance. These instructions, given 24 hours prior to the experiment, included maintaining regular sleep patterns, abstaining from alcohol, and refraining from caffeine intake. Thus, drowsiness in our study was induced gradually by prolonged, monotonous L3 highway driving rather than by sleep restriction.

Upon arrival, participants provided written informed consent and then participated in a 30-minute training session. This session covered the experimental procedure, vehicle operation, cognitive tasks, and subjective questionnaires. Before commencing the experimental drives, physiological sensors were fitted and calibrated. After each drive, participants completed questionnaires to assess their cognitive load using the NASA Task Load Index (NASA-TLX) \cite{hart1988development} (ranging from 1 to 20) and drowsiness level using the Karolinska Sleepiness Scale (KSS) \cite{aakerstedt1990subjective} (ranging from 1 to 10), which served as the ground truth for cognitive load and drowsiness. This study was approved by the Human and Artefacts Research Ethics Committee at the Hong Kong University of Science and Technology (protocol number: HREP-2023-0199).

\subsubsection{Participants}
Before conducting the experiment, a power analysis was performed using MorePower software \cite{campbell2012morepower} to determine the minimum sample size needed. The analysis indicated that a sample size of 24 participants would be sufficient to generate a statistical power of 80\%, with confidence interval = 95\% and an effect size of ($\sigma^2$) of 0.06. 

The study involved a total of 42 participants (25 male, 17 female) with a mean age of 35.28 years (SD = 9.10, range = 23-53 years). Participants were recruited across four age groups (20-60 years) to minimize the influence of age-related factors on physiological signals and enhance the model's generalizability. All participants were required to have a valid driver's license for at least one year and no prior experience with advanced driving systems (ADS). Participants received an hourly compensation of 70 RMB and were given the opportunity to earn a performance-based bonus of up to 30 RMB for completing cognitive tasks.

\subsubsection{Apparatus}
\begin{figure}
\begin{center}
\includegraphics[scale=0.35]{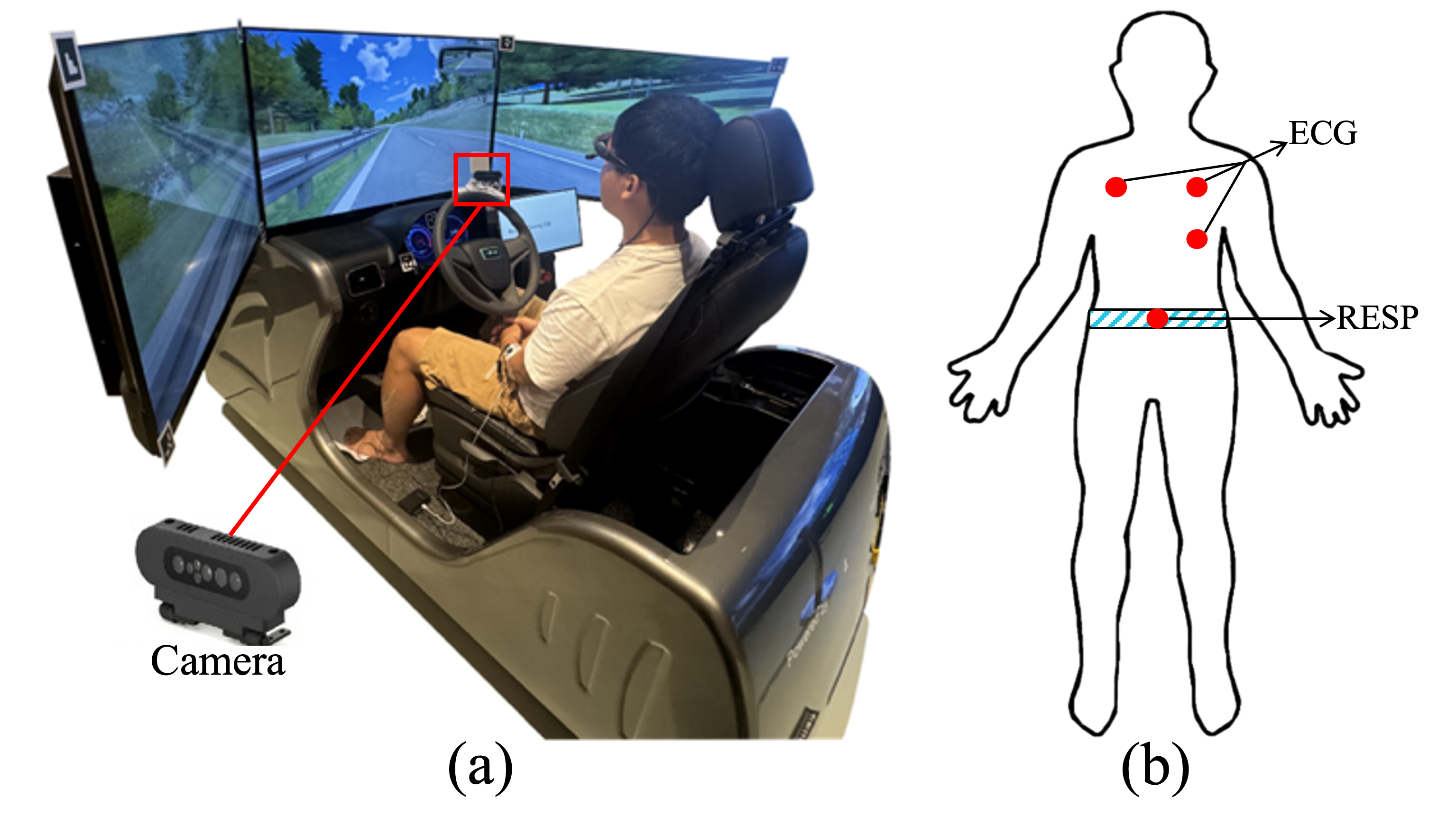}\\
\end{center}
\vspace{-6mm}
\caption{(a) The driving simulator; (b) Data collection sensors.}\label{f2}
\vspace{-4mm}
\end{figure}

The study used a fixed-base driving simulator with three 42-inch screens, providing a 150° horizontal and 47° vertical field of view (Figure \ref{f2}(a)). An external tablet with two touch buttons let participants activate and deactivate the driving automation system. Driving scenarios were developed, and vehicle operation data were recorded at 60 Hz using SILAB 7.1 (WIVW).

\begin{figure}
\begin{center}
\includegraphics[scale=0.09]{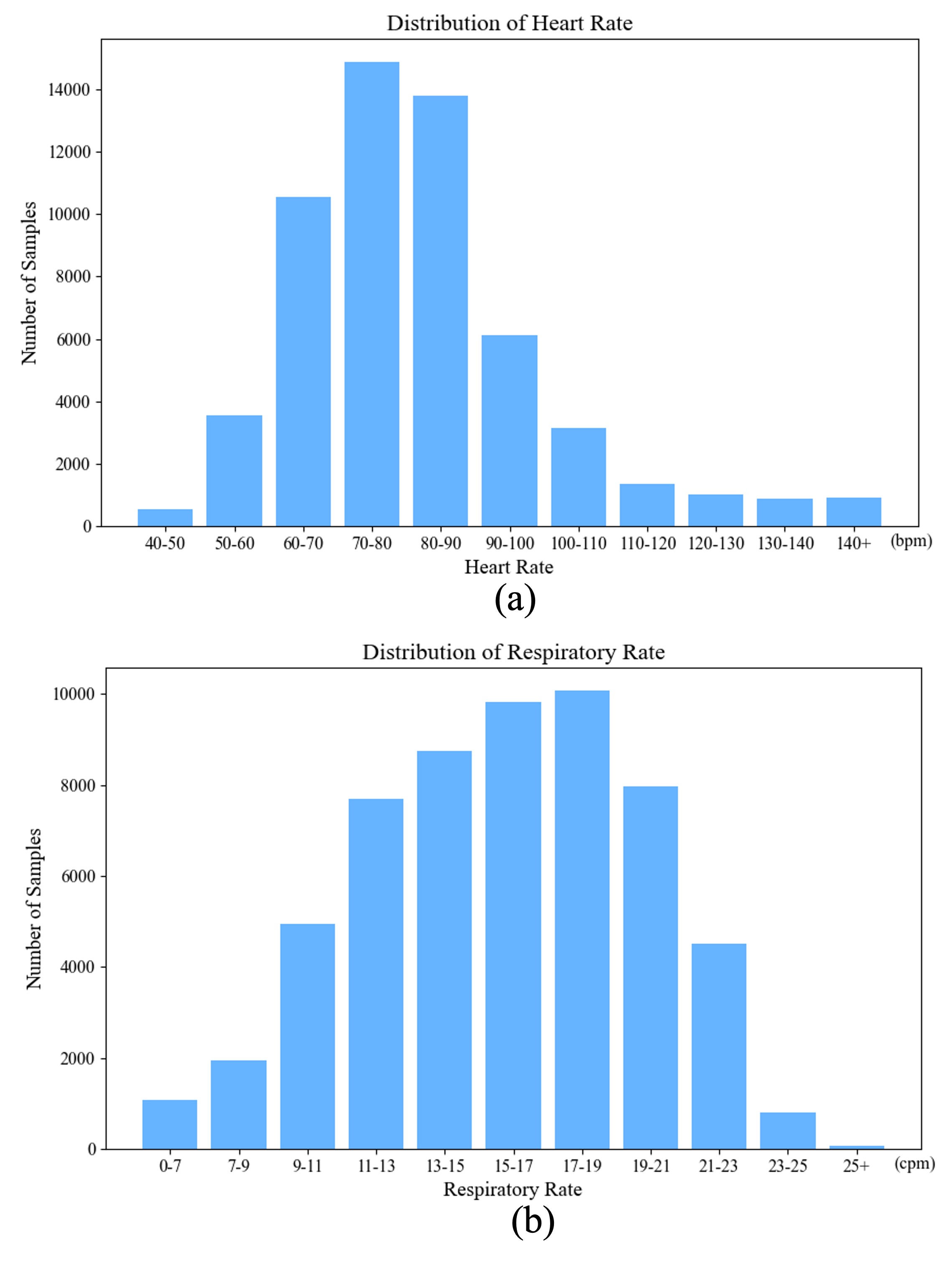}\\
\end{center}
\vspace{-6mm}
\caption{The distribution of HR and RR in MCDD.}\label{f5}
\vspace{-4mm}
\end{figure}

Participants' facial RGB videos were recorded using an Orbbec Gemini Pro camera. The camera was installed in front of the driver, which was recorded at a resolution of 640x480 pixels and a frequency of 30 Hz.  Physiological data, including electrocardiogram (ECG), and respiration (RESP), were collected at a frequency of 100 Hz using a 3-channel SMD electrocardiograph and respiratory belt from Ergoneers (Figure \ref{f2}(b)), based on which the ground truth HR and RR were calculated. In addition to the devices used in this study, participants also wore a lightweight head-mounted eye tracker. However, no eye-tracking signals are used in the present work as VDMoE relies solely on RGB facial video for non-contact driver state and physiological estimation. The distribution of HR and RR in the MCDD dataset is visualized in Figure \ref{f5}.

Besides, following \cite{ahlstrom2024development}, based on the KSS questionnaire collected after each drive, excluding samples with intermediate scores (KSS = 5–7) to avoid ambiguous labels, there are 75.86\% samples labeled as 'awake' (KSS \textless 5), and 24.14\% labeled as `drowsy' (KSS $\geq$ 8). For cognitive workload, we first normalized each participant’s raw NASA-TLX scores to the 1–20 range to mitigate inter-subject differences \cite{yang2021impacts}. We then dichotomized the normalized scores at a global midpoint (10), yielding 29.61\% of drive labeled as `high cognitive load' (score \textgreater 10) and 60.39\% as `normal cognitive load' (score $\leq$10). A sensitivity analysis shifting the threshold by ±2 points showed that the model performance variance was below 5\%, confirming that our median-split strategy produces robust class balances without unduly biasing results. Researchers interested in this dataset can contact the authors to request access.


\subsection{Public Datasets}
We also evaluated our method in two other public datasets. However, as there is no dataset that satisfies the requirements of the multi-task remote detection (i.e., more than one driver state, and contains physiological signals), we selected one dataset for detecting drowsiness (FatigueView) \cite{yang2022FatigueView} and another dataset for multi-task vital signs (V4V) \cite{revanur2021first}.

Specifically, the FatigueView dataset consists of videos captured by RGB and infrared (IR) cameras from five different locations. These videos depict scenarios of drowsy driving and tagged visual cues of drowsiness, varying from subtle to obvious. As no physiological signals or cognitive load levels were collected in this dataset, we utilized the RGB-Front video only from this dataset. In Fatigueview, the drowsiness was labeled by annotators based on drowsiness-related visual signs (e.g., yawning, nodding). In addition, the V4V dataset was widely used in rPPG \cite{wang2023hierarchical, wang2024physmle}. It collected physiological signals and facial videos under ten tasks, and the ground-truth HR and RR were provided in V4V; however, it was not driving-related, and no driver states were assessed.

\begin{table*}[t]
\setlength{\tabcolsep}{1mm}
\centering
\scriptsize
\caption{Performance of Drowsiness and Cognitive Load Estimation on MCDD.}
\begin{tabular}{ccccccccccc}
\toprule 
                          & \multicolumn{4}{c}{\textbf{Drowsiness}}              & \multicolumn{4}{c}{\textbf{Cognitive load}} \\
\cmidrule(lr){2-5} \cmidrule(lr){6-9} 
\textbf{Method} & \textbf{Accuracy↑} & \textbf{F1 Score↑} & \textbf{Sensitivity↑} & \textbf{Specificity↑} & \textbf{Accuracy↑} & \textbf{F1 Score↑} & \textbf{Sensitivity↑} & \textbf{Specificity↑} \\
\midrule 
DBN-HMM$^{+}$ \cite{weng2017driver}                 & 71.96 & 51.57 & 57.25 & 78.63 & — & — & — & — \\
3DCNN+BiLSTM$^{+}$ \cite{yang2020driver}         & 72.18 & 51.09 & 52.84 & 79.86 & — & — & — & — \\
DDDNet$^{+}$ \cite{park2016driver}     & 74.22 & 51.09 & 51.31 & 82.10 & — & — & — & — \\
IsoSSL-MoCo$^{+}$ \cite{mou2021isotropic}             & 76.82 & 56.32 & 59.62 & 87.71 & — & — & — & — \\
ReNeXt3D-101+LSTM$^{+}$ \cite{zhao2020driver}     & 78.61 & 63.30 & 61.32 & 87.45 & — & — & — & — \\
MIGCN$^{+}$ \cite{wei2025prior} & 70.24 & 50.75 & 59.31 & 83.09 & — & — & — & — \\
FacialUnits \cite{cheng2019assessment}     & 73.78 & 52.21 & 53.07 & 79.35 & — & — & — & — \\
2s-STGCN \cite{bai2021two}                & 79.43 & 56.60 & 44.05 & 94.92 & — & — & — & — \\
VBFLFFA \cite{yang2024video}                & 81.05 & 65.39 & 58.81 & 90.78 & — & — & — & — \\
ResNet3D$^{+}$ \cite{hara3dcnns}  & 80.30 & 65.82 & 62.28 & 88.19 & 72.04 & 61.10 & 68.33 & 73.80 \\
ViViT$^{+}$ \cite{arnab2021vivit} & 79.81 & 60.62 & 51.05 & 92.40 & 71.76 & 56.57 & 57.25 & 78.63 \\
\hline
VDMoE                   & \textbf{84.31}$^{*}$ & \textbf{69.84}$^{*}$ & \textbf{66.72}$^{*}$ & \textbf{92.58}$^{*}$ & \textbf{79.96}$^{*}$ & \textbf{68.81}$^{*}$ & \textbf{77.13}$^{*}$ & \textbf{80.77}$^{*}$ \\
\bottomrule 
\end{tabular}
\textnormal{\\Notes: In this and the following tables, the mean of 5-time evaluations is presented. '—' means there are no evaluation results as the model cannot conduct the corresponding task. The \textbf{bold} shows the best result within each column, and the $*$ indicates the significantly (p-value $<$ 0.05 with the paired-t test) best result as compared to the second-best result. $+$ indicates methods taking facial video as models' input.}
\label{t1}
\vspace{-5mm}
\end{table*}

\begin{table}[t]
\setlength{\tabcolsep}{1mm}
\centering
\scriptsize
\caption{Performance of HR and RR Estimation on MCDD.}
\begin{tabular}{cccccccc}
\toprule 
 & \multicolumn{3}{c}{\textbf{HR}}          &        & \multicolumn{3}{c}{\textbf{RR}}                \\
\cmidrule(lr){2-4} \cmidrule(lr){5-8} 
\textbf{Method} & \textbf{MAE↓} & \textbf{RMSE↓} & \textbf{p↑} & & \textbf{MAE↓} & \textbf{RMSE↓} & \textbf{p↑} \\
\midrule 
CHROM$^{+}$ \cite{de2013robust}                  & 18.33 & 19.70 & 0.20 & & — & — & — \\
POS$^{+}$ \cite{wang2016algorithmic}                   & 17.33 & 20.02 & 0.22 & & — & — & — \\
ARM-RR$^{+}$ \cite{tarassenko2014non}                 & — & — & — & & 7.32 & 9.16 & 0.11 \\
Dual-GAN \cite{lu2021dual}                 & 13.29 & 18.86 & 0.31 & & — & — & — \\
ConDiff-rPPG \cite{wang2024diff}                 & 15.32 & 19.58 & 0.29 & & — & — & — \\
PhysFormer++$^{+}$ \cite{yu2023physformer++}             & 13.15 & 18.33 & 0.34 & & — & — & — \\
MTTS-CAN$^{+}$ \cite{liu2020multi}               & 13.96 & 18.31 & 0.30 & & 5.22 & 6.68 & 0.37 \\
BigSmall$^{+}$ \cite{narayanswamy2024bigsmall}              & 13.13 & 18.02 & 0.32 & & 5.26 & 6.72 & 0.37 \\
PhysMLE \cite{wang2024physmle}               & 12.03 & 17.11 & 0.46 & & 5.12 & 7.03 & 0.38 \\
ResNet                & 13.46 & 20.24 & 0.31 & & 5.44 & 7.07 & 0.37 \\
ViT           & 13.00 & 19.69 & 0.35 & & 5.04 & 6.94 & 0.39 \\
ResNet3D$^{+}$ \cite{hara3dcnns}              & 14.27 & 19.02 & 0.29 & & 5.63 & 6.58 & 0.14 \\
ViViT$^{+}$ \cite{arnab2021vivit}                  & 14.92 & 19.61 & 0.28 & & 5.33 & 6.10 & 0.16 \\
\hline
VDMoE                   & \textbf{10.32}$^{*}$ & \textbf{15.37}$^{*}$ & \textbf{0.53}$^{*}$ & & \textbf{4.98}$^{*}$ & \textbf{6.53}$^{*}$ & \textbf{0.45}$^{*}$ \\
\bottomrule 
\end{tabular}
\label{t2}
\vspace{-5mm}
\end{table}

\section{Experiment}
\subsection{Implementation Details}
For the experiments on three datasets, the cross-subject evaluation protocol was utilized. Specifically, for all participants in each dataset, we randomly separated them into training, validation, and test sets with the ratio 6:2:2. Models were trained on the data from participants in the training set, adjusting parameters in the validation set, and only reporting the results in the test set. Before training, since the RGB video was recorded at 30 Hz, to ensure a more accurate RR and HR measurement, we set the sliding window size $F=300$ frames (10 s) and step size $S=30$ (1 s) frames, resulting in a 1-Hz update rate for all model outputs. For each 10-s window, HR and RR are computed from the synchronized ECG and RESP signals on the same temporal window. The drowsiness and cognitive workload labels of each 10-s window are inherited from the KSS and NASA-TLX scores collected at the end of the corresponding drive, under the assumption that the driver’s state remains approximately stationary within such short segments of a 3.8–4.0 minute drive. 

Then, the MediaPipe Face Mesh package from Google\footnote{https://github.com/google-ai-edge/mediapipe} was leveraged to perform landmark detection. Following \cite{yang2024video}, we selected 106 facial landmarks. The size of landmark matrix $I_f$ is $\mathbb{R}^{300\times 106\times 2}$. To standardize the input, each frame of the left eye and right eye $I_l, I_r$ was resized to $\mathbb{R}^{25\times 25\times 3}$ and the mouth frame was resized to $\mathbb{R}^{35\times 15\times 3}$. Additionally, based on 106 landmarks of each sliding window, we further generated STMaps following \cite{niu2019rhythmnet}.  The size of STMap is $\mathbb{R}^{300\times 25\times 3}$. Then, in Table \ref{t_emb}, we present the specific configuration of our feature embedding modules. In practice, we set up $K=4$ experts in each VDMoE block, and there are $L=1$ layers of VDMoE. Adam optimizer with a learning rate of 0.00001 was used for training. The trade-off parameter $k$ was 0.001. The batch size $B$ and maximum iterations were set to 250 and 20000, respectively. 

Following existing works \cite{wang2024multi, peng2024multi}, we used Accuracy, F1 score, Sensitivity, and Specificity to evaluate performance on cognitive load and drowsiness estimation. For HR and RR estimation, mean absolute error (MAE), root mean square error (RMSE), and Pearson’s correlation coefficient (p) were used \cite{wang2024genelizable}. We conducted five evaluations using five different random seeds, and then used a paired t-test to determine the significance of the mean performance difference between the best and second-best models. The whole work was implemented in the Pytorch framework, and all experiments were conducted on an Nvidia RTX A6000.

\begin{table*}[t]
\setlength{\tabcolsep}{1mm}
\centering
\scriptsize
\caption{Ablation Study on Drowsiness, Cognitive Load, HR, and RR Estimation with Computational Cost.}
\begin{tabular}{ccccccccccccc}
\toprule 
                          & \multicolumn{3}{c}{\textbf{Drowsiness}}              & \multicolumn{3}{c}{\textbf{Cognitive load}} & \multicolumn{2}{c}{\textbf{HR}} & \multicolumn{2}{c}{\textbf{RR}} & \multicolumn{2}{c}{\textbf{Computational Cost}} \\
\cmidrule(lr){2-4} \cmidrule(lr){5-7} \cmidrule(lr){8-9} \cmidrule(lr){10-11} \cmidrule(lr){12-13} 
\textbf{Method} & \textbf{Accuracy↑} & \textbf{F1 Score↑} & & \textbf{Accuracy↑} & \textbf{F1 Score↑} & & \textbf{MAE↓} & \textbf{p↑} & \textbf{MAE↓} & \textbf{p↑} & \textbf{Params(M)} & \textbf{FLOPs(G)} \\
\midrule 
ViT        & — & — & & — & — & & 13.00 & 0.35 & 5.34 & 0.27 & 81.82 & 3.41 \\
ResNet                 & — & — & & — & — & & 13.46 & 0.31 & 5.44 & 0.27 & 12.74 & 2.21 \\
ResNet3D$^{+}$ \cite{hara3dcnns}              & 80.30 & 65.82 & & 72.04 & 61.10 & & 14.27 & 0.29 & 5.63 & 0.14 & 31.82 & 40.70 \\
ViViT$^{+}$ \cite{arnab2021vivit}                 & 79.81 & 60.62 & & 71.76 & 56.57 & & 14.92 & 0.20 & 5.33 & 0.16 & 86.57 & 281 \\
VDMoE w/o $I_l$            & 81.65 & 61.44 & & 74.47 & 58.41 & & 11.72 & 0.42 & 5.32 & 0.38 & 6.08 & 1.47 \\
VDMoE w/o $I_r$            & 77.08 & 56.34 & & 73.11 & 61.95 & & 11.90 & 0.45 & 5.15 & 0.43 & 6.08 & 1.47 \\
VDMoE w/o $I_m$            & 79.49 & 60.58 & & 72.42 & 59.01 & & 11.43 & 0.43 & 5.25 & 0.39 & 6.12 & 1.53 \\
VDMoE w/o $I_f$            & 80.69 & 64.17 & & 74.20 & 62.91 & & 11.17 & 0.41 & 5.30 & 0.30 & 6.15 & 1.24 \\
VDMoE w/o $I_s$            & 81.05 & 64.26 & & 74.69 & 63.39 & & 12.72 & 0.40 & 5.13 & 0.41 & 5.81 & 1.64 \\
VDMoE w/o $\mathcal{L}_{align}$         & 82.35 & 67.30 & & 71.22 & 59.12 & & 10.94 & 0.46 & 4.98 & 0.44 & 4.17 & 1.80 \\
\hline
VDMoE                  & \textbf{84.31}$^{*}$ & \textbf{69.84}$^{*}$ & & \textbf{79.96}$^{*}$ & \textbf{68.81}$^{*}$ & & \textbf{10.32}$^{*}$ & \textbf{0.50}$^{*}$ & \textbf{4.98} & \textbf{0.45} & 4.17 & 1.80 \\
\bottomrule 
\end{tabular}
\label{t3}
\vspace{-2mm}
\end{table*}

\subsection{Baseline Models}
To evaluate the performance of our proposed VDMoE, we compared it with several single-task or multi-task models for both state and physiological estimation. Specifically, for single-task models, DBN-HMM \cite{weng2017driver}, 3DCNN+BiLSTM \cite{yang2020driver}, DDDNet \cite{park2016driver}, IsoSSL-MoCo \cite{mou2021isotropic}, ReNeXt3D-101+LSTM \cite{zhao2020driver}, FacialUnits \cite{cheng2019assessment}, 2s-STGCN \cite{bai2021two}, MIGCN \cite{wei2025prior}, and VBFLFF \cite{yang2024video} were used for drowsiness estimation, and CHROM \cite{de2013robust}, POS \cite{wang2016algorithmic}, ARM-RR \cite{tarassenko2014non}, Dual-GAN \cite{lu2021dual}, ConDiff-rPPG \cite{wang2024diff}, and PhysFormer++ \cite{yu2023physformer++} were used for physiological measurement. For multi-task models, since there is currently no facial video-based method proposed for multi-task state and physiological estimation, except for three methods for multi-task physiological estimation (i.e., MTTS-CAN \cite{liu2020multi}, BigSmall \cite{narayanswamy2024bigsmall}, and PhysMLE \cite{wang2024physmle}), we 
built four baselines with the hard-parameter sharing paradigm. In general, each multi-task baseline consists of a typical backbone network (i.e., ResNet3D and ViViT for 3D video input, ResNet and ViT for physiological estimation with STMap input \cite{wang2023hierarchical}) and estimation heads that are the same as VDMoE. 

Among the baselines, following their source papers, DBN-HMM, 3DCNN+BiLSTM, DDDNet, IsoSSL-MoCo, ReNeXt3D-101+LSTM, MIGCN, CHROM, POS, ARM-RR, PhysFormer++, MTTS-CAN, BigSmall, ResNet3D, and ViViT utilize only raw 3D facial video as input. In contrast, FacialUnits and 2s-STGCN rely on facial landmarks extracted from the facial video. Dual-GAN, ConDiff-rPPG, and PhysMLE use STMap as input, while VBFLFF incorporates landmarks along with videos of the eye and mouth subregions. The results are presented in Table \ref{t1}, \ref{t2}.

\subsection{Results of Comparison Experiment}
Firstly, we identified that our proposed VDMoE significantly outperformed all baselines in both state and physiological estimation. Specifically, for single-task state estimation, we found that, methods taking facial video as input generally performed worse than those with preprocessed facial features in drowsiness estimation (e.g., the accuracy of ReNeXt3D-101+LSTM is worse than VBFLFFA by 7.2\%). In addition, two multi-task baselines (i.e., ResNet3D and ViViT) outperformed other single-task methods that used facial videos (e.g., the accuracy of ResNet3D is higher than 3DCNN+BiLSTM by 11.2\%). Nevertheless, our proposed VDMoE still achieved significantly better performance in both drowsiness and cognitive load estimation tasks than the best single-task model (VBFLFFA) by 4\% and the second-best multi-task baseline (ResNet3D) by 5.1\% in accuracy.

Secondly, referring to Table \ref{t2}, VDMoE achieved the best performance on two physiological estimation tasks (HR and RR). In addition, compared to traditional single-task methods (i.e., CHROM, POS, ARM-RR), the rest deep methods have lower estimation errors in both two tasks. Moreover, for four multi-task baselines (i.e., ResNet, ViT, ResNet3D, and ViViT), baselines with facial videos performed worse in HR estimation than those taking preprocessed STMap. For instance, the MAE of ResNet is 5.7\% lower than ResNet3D in HR estimation. Further, in RR estimation task, although baselines with facial videos sometimes had lower estimation error (e.g., the MAE of ViViT is lower than ResNet by 2\%), their p was notably lower than most other deep methods and baselines (e.g., p of ResNet3D is lower than ViT by 64.1\%). 

\subsection{Results of Ablation and Computational Cost Study}
In this part, to elaborate on the effectiveness of VDMoE, we constructed several variants to compare with. In brief, except for four multi-task baselines (i.e., ViT, ResNet, ResNet3D, and ViViT), we constructed six variants by eliminating one facial feature and its corresponding feature embedding module, or the proposed prior-driven alignment regularization $\mathcal{L}_{align}$ from the complete VDMoE. The results are shown in Table \ref{t3}. In addition to multi-task estimation performance, the computational costs (i.e., Parameter number and FLOPs) are also presented.

As shown in Table \ref{t3}, the complete version of VDMoE significantly outperformed all variants without notable computational cost increases. Compared to baselines, VDMoE not only performed better in all tasks, but also with less computational cost (e.g., 23.3\% increase in the MAE of HR estimation but with 69\% parameter size of ResNet).

At the same time, we noticed a notable decrease in the performance of drowsiness estimation (9.4\% in accuracy) when the $I_r$ module was removed from VDMoE, and a decrease in cognitive load estimation when $\mathcal{L}_{align}$ was removed (12\% in accuracy). In addition, for physiological estimation, the variant without $I_s$ has the worst performance in HR estimation, while the lowest estimation precision of RR estimation belongs to the variant without $I_f$. Nevertheless, compared to baselines with facial video, most variants can achieve better multi-task estimation performance, particularly for physiological indicators, and maintain a relatively lower computational cost at the same time. For example, as the variant that performed poorest in HR estimation, the MAE of VDMoE without $I_s$ was still lower than ResNet3D by 10.9\%, even with only around 1/5 parameters. Compared to baselines that can only make physiological estimations (ResNet and ViT), all variants still dominate in the physiological estimation task performance and computational costs.

\begin{figure*}
\begin{center}
\includegraphics[scale=0.06]{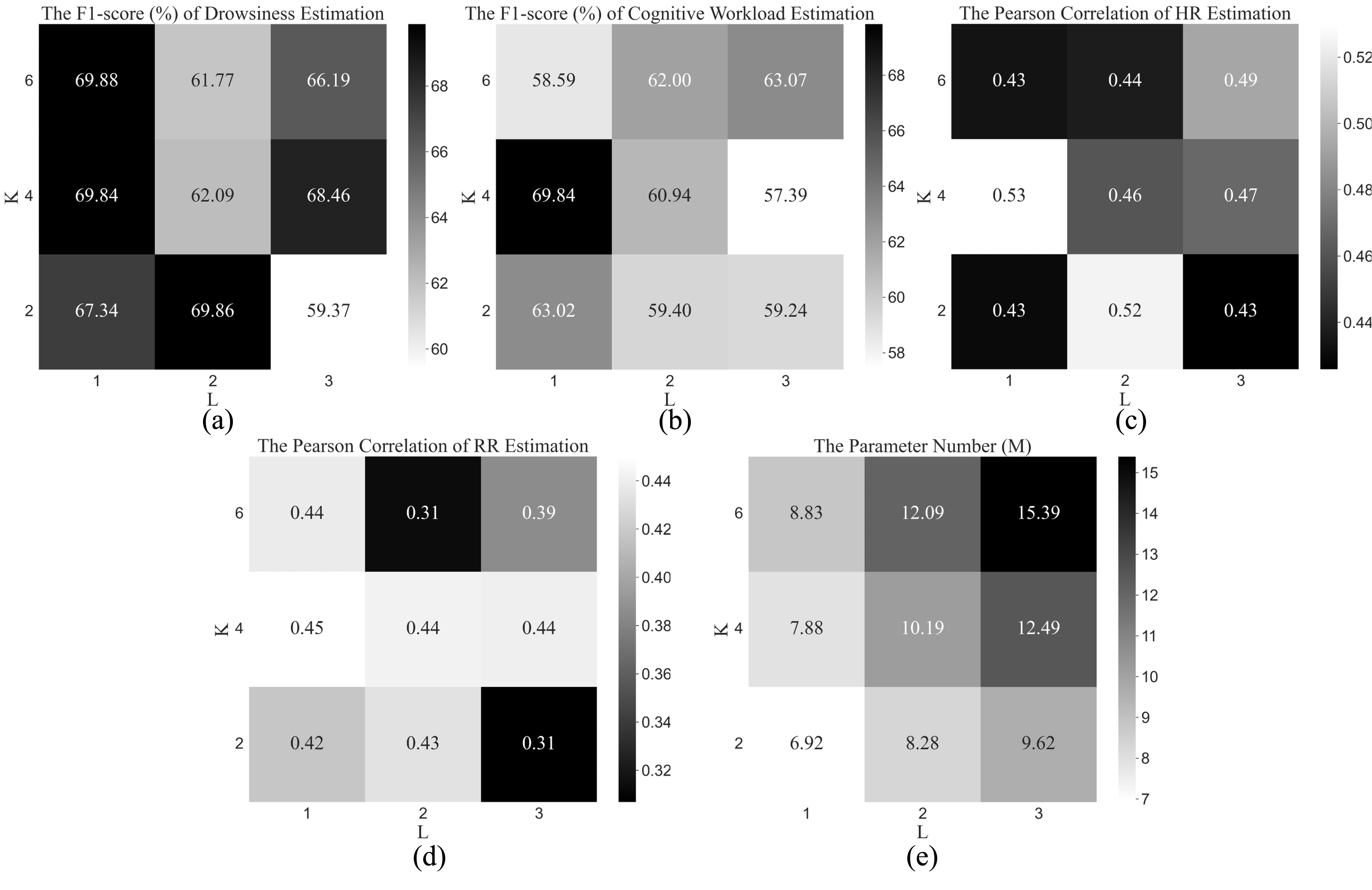}\\
\end{center}
\vspace{-4mm}
\caption{The visualization of the impact of hyper-parameter $K$ and $L$ to each task. Subfigure (a)(b) indicates the F1-score of drowsiness and cognitive load estimation, (b)(d) is the Pearson correlation of HR and RR estimation, and (e) presents the parameter number (M) of VDMoEs with different $K$ and $L$.}\label{f6}
\vspace{-4mm}
\end{figure*}

\subsection{Impact of $K$ and $L$}
The $K$ denotes the number of experts in each layer of the model, while the $L$ denotes the number of layers in the VDMoE. Since the $K$ and $L$ control the width and depth of the VDMoE, we conducted a hyper-parameter test to verify the best choice of $K$ and $L$, and their impact on the multi-task performance. We visualized the results in Figure \ref{f6}.

Firstly, upon examining the drowsiness and cognitive load estimation (Figure \ref{f6}(a)(b)), there did not appear to be a consistent trend with respect to the number of $K$ and $L$; however, it is notable that the highest F1 score for drowsiness and cognitive load detection was both achieved when $K=4$, $L=1$. This suggests that a moderate increase in the number of experts with a single layer is beneficial for these tasks. The Pearson correlation $p$ coefficients in the HR and RR estimation (Figure \ref{f6}(c)(d)) vary across different configurations, with some of the highest values observed with $K=2$, $L=2$ for HR and $K=4$, $L=1$ for RR. It is worth noting that configurations with $K=2$ tend to lead to a higher association for HR, while $K=4$ tends to perform better for RR. 

Additionally, there is a general trend that increasing the number of experts and layers leads to a larger number of parameters, which is consistent with the expectation that more complex models have a higher capacity (and potentially greater computational demands). In general, there is no clear linear correlation between the number of parameters and the performance metrics (F1 scores and Pearson correlations). For instance, the highest number of parameters (15.39 M, see \ref{f6}(e)) does not correspond to the highest performance across tasks, indicating that simply increasing model complexity does not guarantee improved performance.

In conclusion, this analysis demonstrates the trade-off between model complexity and task-specific performance. While no single configuration optimized all metrics simultaneously, the $K=4$, $L=1$ configuration achieved the highest F1-scores for both drowsiness and cognitive workload estimation. Therefore, we selected $K=4$ and $L=1$ for our final VDMoE model, as it provides the most effective and balanced performance across our primary multi-task objectives.

\begin{table}[t]
\setlength{\tabcolsep}{1mm}
\centering
\scriptsize
\caption{Performance of HR and RR Estimation on V4V.}
\begin{tabular}{cccccccc}
\toprule 
 & \multicolumn{3}{c}{\textbf{HR}}          &        & \multicolumn{3}{c}{\textbf{RR}}                \\
\cmidrule(lr){2-4} \cmidrule(lr){5-8} 
\textbf{Method} & \textbf{MAE↓} & \textbf{RMSE↓} & \textbf{p↑} & & \textbf{MAE↓} & \textbf{RMSE↓} & \textbf{p↑} \\
\midrule 
CHROM$^{+}$ \cite{de2013robust}                  & 11.44 & 16.97 & 0.28 & & —     & —     & —     \\
POS$^{+}$ \cite{wang2016algorithmic}                   & 14.59 & 21.26 & 0.19 & & — & — & — \\
ARM-RR$^{+}$ \cite{tarassenko2014non}                 & — & — & — & & 8.11  & 12.23 & 0.08 \\
Dual-GAN \cite{lu2021dual}                  & 4.93  & 7.68  & 0.81 & & — & — & — \\
ConDiff-rPPG \cite{wang2024diff}                  & 5.19  & 7.88  & 0.80 & & — & — & — \\
PhysFormer++$^{+}$ \cite{yu2023physformer++}             & 4.88  & 7.62  & 0.80 & & — & — & — \\
MTTS-CAN$^{+}$ \cite{liu2020multi}               & 5.31  & 8.03  & 0.79 & & 1.98  & 6.31  & 0.15 \\
BigSmall$^{+}$ \cite{narayanswamy2024bigsmall}              & 5.03  & 7.84  & 0.81 & & 1.88  & 5.83  & 0.17 \\
PhysMLE \cite{wang2024physmle}               & 4.79  & 8.06  & 0.79 & & 1.55  & 5.40  & 0.21 \\
ResNet                & 4.64  & 7.98  & 0.80 & & 2.59  & 6.77  & 0.12 \\
ViT           & 4.91  & 8.57  & 0.76 & & 1.94  & 5.98  & 0.17 \\
ResNet3D$^{+}$ \cite{hara3dcnns}              & 8.32  & 11.70 & 0.61  & & 2.03  & 6.09  & 0.09 \\
ViViT$^{+}$ \cite{arnab2021vivit}                  & 7.91  & 9.88  & 0.69 & & 1.81  & 5.89  & 0.16 \\
\hline
VDMoE                   & \textbf{4.01}$^{*}$ & \textbf{6.82}$^{*}$ & \textbf{0.85}$^{*}$ & & \textbf{1.02}$^{*}$ & \textbf{4.94}$^{*}$ & \textbf{0.31}$^{*}$ \\
\bottomrule 
\end{tabular}
\textnormal{\\Notes: In this table, since there are no cognitive load and drowsiness estimation tasks, the VDMoE only applied the two losses for HR and RR regularization.}
\label{t4}
\vspace{-3mm}
\end{table}

\begin{table}[t]
\setlength{\tabcolsep}{1mm}
\centering
\scriptsize
\caption{Performance of Drowsiness Estimation on FatigueView.}
\begin{tabular}{ccccc}
\toprule 
\textbf{Method} & \textbf{Accuracy↑} & \textbf{F1-score↑} & \textbf{Sensitivity↑} & \textbf{Specificity↑}\\
\midrule 
DBN-HMM$^{+}$ \cite{weng2017driver}                 & 79.35 & 81.66 & 86.35 & 80.61 \\
3DCNN+BiLSTM$^{+}$ \cite{yang2020driver}         & 79.21 & 81.03 & 85.40 & 80.27 \\
DDDNet$^{+}$ \cite{park2016driver}     & 83.20 & 86.66 & 89.12 & 83.05 \\
IsoSSL-MoCo$^{+}$ \cite{mou2021isotropic}             & 86.77 & 88.35 & 91.44 & 86.77 \\
ReNeXt3D-101+LSTM$^{+}$ \cite{zhao2020driver}     & 84.32 & 87.60 & 89.33 & 82.22 \\
MIGCN$^{+}$ \cite{wei2025prior} & 80.37 & 82.33 & 88.52 & 82.34 \\
FacialUnits \cite{cheng2019assessment}     & 81.22 & 89.27 & 89.77 & 85.30 \\
2s-STGCN \cite{bai2021two}                & 80.44 & 86.71 & 90.04 & 84.28 \\
VBFLFFA \cite{yang2024video}                & 86.51 & 92.33 & 95.21 & 87.02 \\
ResNet3D$^{+}$ \cite{hara3dcnns} & $\textbf{89.37}^{*}$ & $\textbf{93.78}^{*}$ & 94.79 & $\textbf{91.30}^{*}$ \\
ViViT$^{+}$ \cite{arnab2021vivit} & 82.90 & 90.70 & 83.44 & 88.62 \\
\hline
VDMoE                   & 87.70 & 92.91 & $\textbf{95.03}^{*}$ & 90.40 \\
\bottomrule 
\end{tabular}
\textnormal{\\Notes: In this table, since there is only the drowsiness estimation task, the VDMoE only applied the $\mathcal{L}_{drow}$ loss.}
\label{t5}
\vspace{-5mm}
\end{table}

\subsection{Performance on Other Datasets}
To verify the effectiveness of VDMoE in wider scenarios, we assessed it on two other public datasets (i.e., FatigueView and V4V). The results are presented in Table \ref{t4}, \ref{t5}. For the multi-task remote physiological measurement, the VDMoE still significantly outperformed all methods. For the single-task drowsiness estimation in the FatigueView dataset, despite the VDMoE still outperforming most of the baselines, the ResNet3D stands out with the highest accuracy (89.37\%), F1-score (93.78\%), and specificity (91.30\%). It suggests that the ResNet3D is particularly effective at identifying true negatives, i.e., correctly identifying awake states. Given the much higher computational cost of ResNet3D, VDMoE still demonstrates competitive single-task performance.

\section{Discussion}
Firstly, the VDMoE model's outstanding performance, particularly in comparison with single-task models and multi-task baselines, underscores the effectiveness of its architecture. Notably, the model’s superior accuracy in drowsiness estimation, even when compared to specialized single-task models, suggests that a well-designed multi-task model does not necessarily trade off performance in individual tasks for versatility. Besides, the better performance of VDMoE compared to multi-task baselines without task-specific feature subspace construction (e.g., ResNet3D and ViViT) illustrates that shared low-level information extraction and separate task-related feature space can enhance performance across related tasks by leveraging common features. Moreover, the comparative underperformance of models relying solely on facial video inputs versus those utilizing preprocessed facial features highlights the potential limitations of raw video data in capturing subtle physiological or behavioral cues. This finding suggests that preprocessing steps or feature engineering might still play a crucial role in optimizing model performance, even in an era where end-to-end deep learning models are prevalent. Particularly, considering both facial features at the action level (e.g., blinks and expressions) and changes in color (STMap) during pre-processing may help extract changes in physiological indicators associated with the driver states.

In addition, the inclusion of a prior-driven loss, denoted as $\mathcal{L}_{align}$ plays a pivotal role in aligning the model's outputs with prior knowledge. This loss function helps guide the learning process, which ensures that the model not only optimizes estimation accuracy but also considers the underlying structure or relationships within the data that are known a priori. By incorporating $\mathcal{L}_{align}$, the model is encouraged to learn representations that are consistent with the findings in previous research (i.e., the correlation between drowsiness and cognitive load), potentially leading to more robust and generalizable performance. In the future, exploring more sophisticated forms of prior-driven losses that can be explicitly guided by subtle relationships among tasks could further improve the model's robustness and generalizability.

Lastly, from a deployment perspective, VDMoE is designed to be lightweight and compatible with real-time driver monitoring. As summarized in Table VII, our model contains only 4.17M parameters and requires 1.80 GFLOPs per 10-s input window, which is an order of magnitude lower than 3D CNN and Transformer baselines (e.g., ResNet3D: 31.82M parameters and 40.70 GFLOPs; ViViT: 86.57M parameters and 281 GFLOPs). This computational footprint is well within the capability of modern automotive-grade GPUs and System-on-Chips. The choice to structure the VDMoE as a multi-task MoE model using MLPs, avoiding more complex networks, is to ensure both the model's performance and its computational efficiency. While 3DCNNs and Transformers have demonstrated remarkable capabilities in capturing spatial and temporal dependencies in data, they come with significantly higher computational costs and parameter counts. For real-time applications, such as driver state monitoring systems, where low latency is crucial, the efficiency of the model is as important as its accuracy. MLPs, by contrast, offer a simpler and more computationally efficient alternative. When structured as part of an MoE framework, MLPs can be highly effective in capturing complex relationships within the data. Each "expert" in the MoE model can specialize in different aspects or features of the data, and the "gating" mechanism can learn to dynamically allocate computational resources by activating relevant experts for a given task or input. This allows the VDMoE to maintain a balance between model complexity and computational efficiency, making it suitable for deployment in real-world settings where resources may be constrained. Future studies should investigate the extension of the MoE framework to incorporate other types of network architectures, such as lightweight CNNs or compact Transformer variants, which could offer a pathway to further decrease computational demands in the deployment.

\section{Limitations}
While the VDMoE model shows strong performance in multi-task driver state and physiological measurement estimation for conditional autonomous driving, several limitations remain. First, the MCDD dataset is collected in a fixed-base simulator and assumes that the driver’s face is mostly visible and reasonably well illuminated. In real SAE L3 driving with extreme head poses, substantial occlusions, or strong illumination changes, the quality of facial and rPPG features will likely degrade, reducing model performance. The STMap-based rPPG pipeline in VDMoE (including RGB-to-YUV conversion and band-pass filtering) helps mitigate slow global illumination changes and improves signal-to-noise ratio (SNR) \cite{yu2023physformer++,kim2022study}, but severe local shadows, specular reflections, and large head or body movements can still substantially degrade physiological signals. Real vehicles also introduce camera motion and vibration. Prior work on camera-based RR estimation in vehicles shows that methods can be accurate when the vehicle is stationary or moving slowly, but accuracy drops notably at higher speeds \cite{othman2022contactless}. Given the high variability of real driving, relying on a single RGB camera may limit the reliability of physiological and state estimation.

Second, MCDD participants were recruited from a single country/region, with relatively homogeneous cultural backgrounds and similar skin tones. Currently, no multi-task dataset from other regions supports all four tasks with facial video, preventing systematic evaluation of cross-region and cross-demographic generalization. The model’s ability to adapt to such variations in real time, particularly without retraining or manual recalibration, therefore remains unclear. Moreover, there is no on-road dataset that captures multiple driver states and real-world challenges, including but not limited to varying illumination, motion, and user characteristics. Thus, it is critical to collect new on-road datasets and further evaluate and refine our models that worked well on MCDD.

Third, this study targets segment-level (10-s) detection of drowsiness and cognitive workload, assuming that these states vary slowly over each 3.8–4.0 min drive. Although prolonged monotonous L3 driving (1.5 h) is theoretically sufficient to induce drowsiness, our protocol lacks dense subjective ratings, preventing precise estimation of onset latency and fine-grained temporal dynamics within each drive. Future work should develop models that explicitly capture the temporal evolution of drowsiness and time-to-drowsiness.

Lastly, we discretize cognitive load levels using a custom scheme derived from the NASA-TLX questionnaire. While this method is reasonably robust, it inevitably causes information loss. Meanwhile, the individual biases inherent in subjective questionnaires may also jeopardize the model’s performance in estimating drivers’ true states in real-world applications. Future work should explore strategies for more accurate driver state categorization and overcome the limitations of subjective driver state labels.

\section{Conclusion}

We introduced VDMoE, an innovative model for driver-state and physiological measurement estimation that leverages facial video cues. Our method combines an MoE framework with a lightweight MLP, avoiding the computational cost of 3D CNNs and Transformers. VDMoE distills key facial information and uses a prior-driven loss, $\mathcal{L}_{\text{align}}$, to align predictions with known physiological and behavioral patterns, improving accuracy and generalization. This design captures temporal and spatial dynamics, achieving strong performance in drowsiness, cognitive load, HR, and RR estimation.

VDMoE is intended as a core component of real-time DMS in SAE Level-2/3 driving. Its lightweight, video-based design enables direct integration into in-vehicle systems, processing RGB streams from a cabin-facing camera to infer cognitive load and drowsiness. The model’s multi-task outputs can be fused with vehicle control modules or human–machine interfaces (HMI) to trigger adaptive alerts, issue take-over requests, or log physiological states for long-term analysis, highlighting its potential for future intelligent cockpit systems. We also introduced MCDD, a large-scale dataset for monitoring driver states in conditional autonomous driving using RGB video. Collected on a driving simulator with strict experimental protocols, MCDD captures the co-occurrence of multi-dimensional cognitive load and drowsiness in realistic scenarios. We expect it to support future research, particularly in addressing real-world challenges.

Future work includes designing additional multi-task datasets and validating VDMoE in real-world conditional autonomous driving, especially under extreme conditions with very high cognitive load. We also plan to refine the alignment between drowsiness and cognitive workload distributions by integrating human factors knowledge into data-driven models, and to deploy VDMoE on real-time embedded platforms within existing DMS pipelines.

\bibliographystyle{IEEEtran}
\bibliography{IEEEabrv,ref}

\begin{thebibliography}{10}
\providecommand{\url}[1]{#1}
\csname url@samestyle\endcsname
\providecommand{\newblock}{\relax}
\providecommand{\bibinfo}[2]{#2}
\providecommand{\BIBentrySTDinterwordspacing}{\spaceskip=0pt\relax}
\providecommand{\BIBentryALTinterwordstretchfactor}{4}
\providecommand{\BIBentryALTinterwordspacing}{\spaceskip=\fontdimen2\font plus
\BIBentryALTinterwordstretchfactor\fontdimen3\font minus \fontdimen4\font\relax}
\providecommand{\BIBforeignlanguage}[2]{{%
\expandafter\ifx\csname l@#1\endcsname\relax
\typeout{** WARNING: IEEEtran.bst: No hyphenation pattern has been}%
\typeout{** loaded for the language `#1'. Using the pattern for}%
\typeout{** the default language instead.}%
\else
\language=\csname l@#1\endcsname
\fi
#2}}
\providecommand{\BIBdecl}{\relax}
\BIBdecl

\bibitem{kaplan2015driver}
S.~Kaplan, M.~A. Guvensan, A.~G. Yavuz, and Y.~Karalurt, ``Driver behavior analysis for safe driving: A survey,'' \emph{IEEE Transactions on Intelligent Transportation Systems}, vol.~16, no.~6, pp. 3017--3032, 2015.

\bibitem{peng2018vehicle}
Z.~Peng, S.~Gao, Z.~Li, B.~Xiao, and Y.~Qian, ``Vehicle safety improvement through deep learning and mobile sensing,'' \emph{IEEE network}, vol.~32, no.~4, pp. 28--33, 2018.

\bibitem{who2023global}
{World Health Organization}, ``Global status report on road safety 2023,'' World Health Organization, Geneva, Tech. Rep., 2023.

\bibitem{sae2018taxonomy}
S.~International, ``Taxonomy and definitions for terms related to driving automation systems for on-road motor vehicles,'' \emph{SAE international}, vol. 4970, no. 724, pp. 1--5, 2018.

\bibitem{baumler1994validity}
G.~B{\"a}umler, ``On the validity of the yerkes-dodson law,'' \emph{Studia Psychologica}, vol.~36, no.~3, p. 205, 1994.

\bibitem{perello2021driver}
J.~R. Perello-March, C.~G. Burns, R.~Woodman, M.~T. Elliott, and S.~A. Birrell, ``Driver state monitoring: Manipulating reliability expectations in simulated automated driving scenarios,'' \emph{IEEE transactions on intelligent transportation systems}, vol.~23, no.~6, pp. 5187--5197, 2021.

\bibitem{recarte2000effects}
M.~A. Recarte and L.~M. Nunes, ``Effects of verbal and spatial-imagery tasks on eye fixations while driving.'' \emph{Journal of experimental psychology: Applied}, vol.~6, no.~1, p.~31, 2000.

\bibitem{muhrer2011effect}
E.~Muhrer and M.~Vollrath, ``The effect of visual and cognitive distraction on driver’s anticipation in a simulated car following scenario,'' \emph{Transportation research part F: traffic psychology and behaviour}, vol.~14, no.~6, pp. 555--566, 2011.

\bibitem{soares2020analyzing}
S.~Soares, T.~Monteiro, A.~Lobo, A.~Couto, L.~Cunha, and S.~Ferreira, ``Analyzing driver drowsiness: From causes to effects,'' \emph{Sustainability}, vol.~12, no.~5, p. 1971, 2020.

\bibitem{he2019high}
D.~He, B.~Donmez, C.~C. Liu, and K.~N. Plataniotis, ``High cognitive load assessment in drivers through wireless electroencephalography and the validation of a modified n-back task,'' \emph{IEEE Transactions on Human-Machine Systems}, vol.~49, no.~4, pp. 362--371, 2019.

\bibitem{doudou2020driver}
M.~Doudou, A.~Bouabdallah, and V.~Berge-Cherfaoui, ``Driver drowsiness measurement technologies: Current research, market solutions, and challenges,'' \emph{International Journal of Intelligent Transportation Systems Research}, vol.~18, pp. 297--319, 2020.

\bibitem{meteier2021classification}
Q.~Meteier, M.~Capallera, S.~Ruffieux, L.~Angelini, O.~Abou~Khaled, E.~Mugellini, M.~Widmer, and A.~Sonderegger, ``Classification of drivers' workload using physiological signals in conditional automation,'' \emph{Frontiers in psychology}, vol.~12, p. 596038, 2021.

\bibitem{qu2023driver}
Y.~Qu, H.~Hu, J.~Liu, Z.~Zhang, Y.~Li, and X.~Ge, ``Driver state monitoring technology for conditionally automated vehicles: Review and future prospects,'' \emph{IEEE Transactions on Instrumentation and Measurement}, 2023.

\bibitem{hu2024eeg}
F.~Hu, L.~Zhang, X.~Yang, and W.-A. Zhang, ``Eeg-based driver fatigue detection using spatio-temporal fusion network with brain region partitioning strategy,'' \emph{IEEE Transactions on Intelligent Transportation Systems}, 2024.

\bibitem{he2022classification}
D.~He, Z.~Wang, E.~B. Khalil, B.~Donmez, G.~Qiao, and S.~Kumar, ``Classification of driver cognitive load: Exploring the benefits of fusing eye-tracking and physiological measures,'' \emph{Transportation research record}, vol. 2676, no.~10, pp. 670--681, 2022.

\bibitem{picot2011line}
A.~Picot, S.~Charbonnier, and A.~Caplier, ``On-line detection of drowsiness using brain and visual information,'' \emph{IEEE Transactions on systems, man, and cybernetics-part A: systems and humans}, vol.~42, no.~3, pp. 764--775, 2011.

\bibitem{li2015smartwatch}
G.~Li, B.-L. Lee, and W.-Y. Chung, ``Smartwatch-based wearable eeg system for driver drowsiness detection,'' \emph{IEEE Sensors Journal}, vol.~15, no.~12, pp. 7169--7180, 2015.

\bibitem{murugan2020detection}
S.~Murugan, J.~Selvaraj, and A.~Sahayadhas, ``Detection and analysis: Driver state with electrocardiogram (ecg),'' \emph{Physical and engineering sciences in medicine}, vol.~43, no.~2, pp. 525--537, 2020.

\bibitem{yang2024video}
L.~Yang, H.~Yang, H.~Wei, Z.~Hu, and C.~Lv, ``Video-based driver drowsiness detection with optimised utilization of key facial features,'' \emph{IEEE Transactions on Intelligent Transportation Systems}, 2024.

\bibitem{sikander2020novel}
G.~Sikander and S.~Anwar, ``A novel machine vision-based 3d facial action unit identification for fatigue detection,'' \emph{IEEE Transactions on Intelligent Transportation Systems}, vol.~22, no.~5, pp. 2730--2740, 2020.

\bibitem{mou2021isotropic}
L.~Mou, C.~Zhou, P.~Xie, P.~Zhao, R.~Jain, W.~Gao, and B.~Yin, ``Isotropic self-supervised learning for driver drowsiness detection with attention-based multimodal fusion,'' \emph{IEEE Transactions on Multimedia}, vol.~25, pp. 529--542, 2021.

\bibitem{peng2024multi}
Y.~Peng, H.~Deng, G.~Xiang, X.~Wu, X.~Yu, Y.~Li, and T.~Yu, ``A multi-source fusion approach for driver fatigue detection using physiological signals and facial image,'' \emph{IEEE Transactions on Intelligent Transportation Systems}, 2024.

\bibitem{huang2024self}
Y.~Huang, C.~Liu, F.~Chang, and Y.~Lu, ``Self-supervised multi-granularity graph attention network for vision-based driver fatigue detection,'' \emph{IEEE Transactions on Emerging Topics in Computational Intelligence}, 2024.

\bibitem{huynh2017detection}
X.-P. Huynh, S.-M. Park, and Y.-G. Kim, ``Detection of driver drowsiness using 3d deep neural network and semi-supervised gradient boosting machine,'' in \emph{Computer Vision--ACCV 2016 Workshops: ACCV 2016 International Workshops, Taipei, Taiwan, November 20-24, 2016, Revised Selected Papers, Part III 13}.\hskip 1em plus 0.5em minus 0.4em\relax Springer, 2017, pp. 134--145.

\bibitem{du2020vision}
G.~Du, T.~Li, C.~Li, P.~X. Liu, and D.~Li, ``Vision-based fatigue driving recognition method integrating heart rate and facial features,'' \emph{IEEE transactions on intelligent transportation systems}, vol.~22, no.~5, pp. 3089--3100, 2020.

\bibitem{hasan2024vision}
M.~Z. Hasan, J.~Chen, J.~Wang, M.~S. Rahman, A.~Joshi, S.~Velipasalar, C.~Hegde, A.~Sharma, and S.~Sarkar, ``Vision-language models can identify distracted driver behavior from naturalistic videos,'' \emph{IEEE Transactions on Intelligent Transportation Systems}, 2024.

\bibitem{wang2024revisiting}
J.~Wang, A.~Wang, S.~Yan, D.~He, and K.~Wu, ``Revisiting interactions of multiple driver states in heterogenous population and cognitive tasks,'' \emph{arXiv preprint arXiv:2412.13574}, 2024.

\bibitem{ayas2023drowsiness}
S.~Ayas, B.~Donmez, and X.~Tang, ``Drowsiness mitigation through driver state monitoring systems: a scoping review,'' \emph{Human factors}, p. 00187208231208523, 2023.

\bibitem{wang2024association}
A.~Wang, C.~Huang, J.~Wang, and D.~He, ``The association between physiological and eye-tracking metrics and cognitive load in drivers: A meta-analysis,'' \emph{Transportation Research Part F: Traffic Psychology and Behaviour}, vol. 104, pp. 474--487, 2024.

\bibitem{wang2024physmle}
J.~Wang, H.~Lu, A.~Wang, X.~Yang, Y.~Chen, D.~He, and K.~Wu, ``Physmle: Generalizable and priors-inclusive multi-task remote physiological measurement,'' \emph{IEEE Transactions on Pattern Analysis and Machine Intelligence}, 2025.

\bibitem{massoz2016ulg}
Q.~Massoz, T.~Langohr, C.~Fran{\c{c}}ois, and J.~G. Verly, ``The ulg multimodality drowsiness database (called drozy) and examples of use,'' in \emph{2016 IEEE Winter Conference on Applications of Computer Vision (WACV)}.\hskip 1em plus 0.5em minus 0.4em\relax IEEE, 2016, pp. 1--7.

\bibitem{weng2017driver}
C.-H. Weng, Y.-H. Lai, and S.-H. Lai, ``Driver drowsiness detection via a hierarchical temporal deep belief network,'' in \emph{Computer Vision--ACCV 2016 Workshops: ACCV 2016 International Workshops, Taipei, Taiwan, November 20-24, 2016, Revised Selected Papers, Part III 13}.\hskip 1em plus 0.5em minus 0.4em\relax Springer, 2017, pp. 117--133.

\bibitem{ghoddoosian2019realistic}
R.~Ghoddoosian, M.~Galib, and V.~Athitsos, ``A realistic dataset and baseline temporal model for early drowsiness detection,'' in \emph{Proceedings of the IEEE Conference on Computer Vision and Pattern Recognition Workshops}, 2019, pp. 0--0.

\bibitem{jOrtega2020}
J.~D. Ortega, N.~Kose, P.~Cañas, M.-A. Chao, A.~Unnervik, M.~Nieto, O.~Otaegui, and L.~Salgado, ``Dmd: A large-scale multi-modal driver monitoring dataset for attention and alertness analysis,'' in \emph{Computer Vision -- ECCV 2020 Workshops}, A.~Bartoli and A.~Fusiello, Eds.\hskip 1em plus 0.5em minus 0.4em\relax Springer International Publishing, 2020, pp. 387--405.

\bibitem{yang2022FatigueView}
C.~Yang, Z.~Yang, W.~Li, and J.~See, ``Fatigueview: A multi-camera video dataset for vision-based drowsiness detection,'' \emph{IEEE Transactions on Intelligent Transportation Systems}, vol.~24, no.~1, pp. 233--246, 2022.

\bibitem{jo2024mocas}
W.~Jo, R.~Wang, G.-E. Cha, S.~Sun, R.~K. Senthilkumaran, D.~Foti, and B.-C. Min, ``Mocas: A multimodal dataset for objective cognitive workload assessment on simultaneous tasks,'' \emph{IEEE Transactions on Affective Computing}, 2024.

\bibitem{yang2023real}
H.~Yang, J.~Wu, Z.~Hu, and C.~Lv, ``Real-time driver cognitive workload recognition: Attention-enabled learning with multimodal information fusion,'' \emph{IEEE Transactions on Industrial Electronics}, vol.~71, no.~5, pp. 4999--5009, 2023.

\bibitem{verkruysse2008remote}
W.~Verkruysse, L.~O. Svaasand, and J.~S. Nelson, ``Remote plethysmographic imaging using ambient light.'' \emph{Optics express}, vol.~16, no.~26, pp. 21\,434--21\,445, 2008.

\bibitem{shazeer2016outrageously}
N.~Shazeer, A.~Mirhoseini, K.~Maziarz, A.~Davis, Q.~Le, G.~Hinton, and J.~Dean, ``Outrageously large neural networks: The sparsely-gated mixture-of-experts layer,'' in \emph{International Conference on Learning Representations}, 2016.

\bibitem{hunter2012selection}
D.~Hunter, H.~Yu, M.~S. Pukish~III, J.~Kolbusz, and B.~M. Wilamowski, ``Selection of proper neural network sizes and architectures—a comparative study,'' \emph{IEEE Transactions on Industrial Informatics}, vol.~8, no.~2, pp. 228--240, 2012.

\bibitem{he2016deep}
K.~He, X.~Zhang, S.~Ren, and J.~Sun, ``Deep residual learning for image recognition,'' in \emph{Proceedings of the IEEE conference on computer vision and pattern recognition}, 2016, pp. 770--778.

\bibitem{vaswani2017attention}
A.~Vaswani, N.~Shazeer, N.~Parmar, J.~Uszkoreit, L.~Jones, A.~N. Gomez, {\L}.~Kaiser, and I.~Polosukhin, ``Attention is all you need,'' \emph{Advances in neural information processing systems}, vol.~30, 2017.

\bibitem{saxby2013active}
D.~J. Saxby, G.~Matthews, J.~S. Warm, E.~M. Hitchcock, and C.~Neubauer, ``Active and passive fatigue in simulated driving: discriminating styles of workload regulation and their safety impacts.'' \emph{Journal of experimental psychology: applied}, vol.~19, no.~4, p. 287, 2013.

\bibitem{wang2025towards}
J.~Wang, S.~Ayas, J.~Zhang, X.~Wen, D.~He, and B.~Donmez, ``Towards generalizable drowsiness monitoring with physiological sensors: A preliminary study,'' in \emph{Proceedings of the Human Factors and Ergonomics Society Annual Meeting}, vol.~69, no.~1.\hskip 1em plus 0.5em minus 0.4em\relax SAGE Publications Sage CA: Los Angeles, CA, 2025, pp. 1145--1150.

\bibitem{deng2024analysis}
M.~Deng, A.~Gluck, Y.~Zhao, D.~Li, C.~C. Menassa, V.~R. Kamat, and J.~Brinkley, ``An analysis of physiological responses as indicators of driver takeover readiness in conditionally automated driving,'' \emph{Accident Analysis \& Prevention}, vol. 195, p. 107372, 2024.

\bibitem{lohani2019review}
M.~Lohani, B.~R. Payne, and D.~L. Strayer, ``A review of psychophysiological measures to assess cognitive states in real-world driving,'' \emph{Frontiers in human neuroscience}, vol.~13, p.~57, 2019.

\bibitem{wang2024cognitive}
A.~Wang, J.~Wang, W.~Shi, and D.~He, ``Cognitive workload estimation in conditionally automated vehicles using transformer networks based on physiological signals,'' \emph{Transportation Research Record}, p. 03611981241250023, 2024.

\bibitem{cui2022eeg}
J.~Cui, Z.~Lan, O.~Sourina, and W.~M{\"u}ller-Wittig, ``Eeg-based cross-subject driver drowsiness recognition with an interpretable convolutional neural network,'' \emph{IEEE Transactions on Neural Networks and Learning Systems}, vol.~34, no.~10, pp. 7921--7933, 2022.

\bibitem{wang2026drowsydg}
J.~Wang, W.~Li, Z.~Wang, S.~Ayas, B.~Donmez, D.~He, and K.~Wu, ``Drowsydg-phys: Generalizable driver drowsiness estimation in conditional automated vehicles using physiological signals,'' \emph{Accident Analysis \& Prevention}, vol. 228, p. 108407, 2026.

\bibitem{cella2010measuring}
M.~Cella and T.~Chalder, ``Measuring fatigue in clinical and community settings,'' \emph{Journal of psychosomatic research}, vol.~69, no.~1, pp. 17--22, 2010.

\bibitem{de2013robust}
G.~De~Haan and V.~Jeanne, ``Robust pulse rate from chrominance-based rppg,'' \emph{IEEE Transactions on Biomedical Engineering}, vol.~60, no.~10, pp. 2878--2886, 2013.

\bibitem{wang2016algorithmic}
W.~Wang, A.~C. Den~Brinker, S.~Stuijk, and G.~De~Haan, ``Algorithmic principles of remote ppg,'' \emph{IEEE Transactions on Biomedical Engineering}, vol.~64, no.~7, pp. 1479--1491, 2016.

\bibitem{tarassenko2014non}
L.~Tarassenko, M.~Villarroel, A.~Guazzi, J.~Jorge, D.~Clifton, and C.~Pugh, ``Non-contact video-based vital sign monitoring using ambient light and auto-regressive models,'' \emph{Physiological measurement}, vol.~35, no.~5, p. 807, 2014.

\bibitem{lu2021dual}
H.~Lu, H.~Han, and S.~K. Zhou, ``Dual-gan: Joint bvp and noise modeling for remote physiological measurement,'' in \emph{Proceedings of the IEEE/CVF Conference on Computer Vision and Pattern Recognition}, 2021, pp. 12\,404--12\,413.

\bibitem{yu2023physformer++}
Z.~Yu, Y.~Shen, J.~Shi, H.~Zhao, Y.~Cui, J.~Zhang, P.~Torr, and G.~Zhao, ``Physformer++: Facial video-based physiological measurement with slowfast temporal difference transformer,'' \emph{International Journal of Computer Vision}, vol. 131, no.~6, pp. 1307--1330, 2023.

\bibitem{liu2020multi}
X.~Liu, J.~Fromm, S.~Patel, and D.~McDuff, ``Multi-task temporal shift attention networks for on-device contactless vitals measurement,'' \emph{Advances in Neural Information Processing Systems}, vol.~33, pp. 19\,400--19\,411, 2020.

\bibitem{narayanswamy2024bigsmall}
G.~Narayanswamy, Y.~Liu, Y.~Yang, C.~Ma, X.~Liu, D.~McDuff, and S.~Patel, ``Bigsmall: Efficient multi-task learning for disparate spatial and temporal physiological measurements,'' in \emph{Proceedings of the IEEE/CVF Winter Conference on Applications of Computer Vision}, 2024, pp. 7914--7924.

\bibitem{wang2025align}
J.~Wang, X.~Yang, H.~Lu, D.~He, and K.~Wu, ``Align the gap: Prior-based unified multi-task remote physiological measurement framework for domain generalization and personalization,'' \emph{arXiv preprint arXiv:2506.16160}, 2025.

\bibitem{bai2021two}
J.~Bai, W.~Yu, Z.~Xiao, V.~Havyarimana, A.~C. Regan, H.~Jiang, and L.~Jiao, ``Two-stream spatial--temporal graph convolutional networks for driver drowsiness detection,'' \emph{IEEE Transactions on Cybernetics}, vol.~52, no.~12, pp. 13\,821--13\,833, 2021.

\bibitem{su2018yawning}
M.~Su-Gang, Z.~Chen, S.~Han-Lin, and H.~Jungang, ``Yawning detection algorithm based on convolutional neural network,'' \emph{Comput. Sci}, 2018.

\bibitem{yang2020driver}
H.~Yang, L.~Liu, W.~Min, X.~Yang, and X.~Xiong, ``Driver yawning detection based on subtle facial action recognition,'' \emph{IEEE Transactions on Multimedia}, vol.~23, pp. 572--583, 2020.

\bibitem{park2016driver}
S.~Park, F.~Pan, S.~Kang, and C.~D. Yoo, ``Driver drowsiness detection system based on feature representation learning using various deep networks,'' in \emph{Asian conference on computer vision}.\hskip 1em plus 0.5em minus 0.4em\relax Springer, 2016, pp. 154--164.

\bibitem{zhao2020driver}
L.~Zhao, Z.~Wang, G.~Zhang, and H.~Gao, ``Driver drowsiness recognition via transferred deep 3d convolutional network and state probability vector,'' \emph{Multimedia Tools and Applications}, vol.~79, no.~35, pp. 26\,683--26\,701, 2020.

\bibitem{cheng2019assessment}
Q.~Cheng, W.~Wang, X.~Jiang, S.~Hou, and Y.~Qin, ``Assessment of driver mental fatigue using facial landmarks,'' \emph{IEEE Access}, vol.~7, pp. 150\,423--150\,434, 2019.

\bibitem{wang2023hierarchical}
J.~Wang, H.~Lu, A.~Wang, Y.~Chen, and D.~He, ``Hierarchical style-aware domain generalization for remote physiological measurement,'' \emph{IEEE Journal of Biomedical and Health Informatics}, vol.~28, no.~3, pp. 1635--1643, 2024.

\bibitem{wang2024diff}
J.~Wang, X.~Wei, H.~Lu, Y.~Chen, and D.~He, ``Condiff-rppg: Robust remote physiological measurement to heterogeneous occlusions,'' \emph{IEEE Journal of Biomedical and Health Informatics}, pp. 1--13, 2024.

\bibitem{yang2025consistencyenhancetesttimeadaptation}
\BIBentryALTinterwordspacing
X.~Yang, Y.~Fan, C.~Liu, H.~Su, W.~Guo, J.~Wang, and D.~He, ``Not only consistency: Enhance test-time adaptation with spatio-temporal inconsistency for remote physiological measurement,'' 2025. [Online]. Available: \url{https://arxiv.org/abs/2507.07908}
\BIBentrySTDinterwordspacing

\bibitem{niu2018automatic}
X.~Niu, H.~Han, J.~Zeng, X.~Sun, S.~Shan, Y.~Huang, S.~Yang, and X.~Chen, ``Automatic engagement prediction with gap feature,'' in \emph{Proceedings of the 20th ACM International Conference on Multimodal Interaction}, 2018, pp. 599--603.

\bibitem{wang2025physdrive}
J.~Wang, X.~Yang, Q.~Hu, J.~Tang, C.~Liu, D.~He, Y.~Wang, Y.~Chen, and K.~Wu, ``Physdrive: A multimodal remote physiological measurement dataset for in-vehicle driver monitoring,'' \emph{arXiv preprint arXiv:2507.19172}, 2025.

\bibitem{riquelme2021scaling}
C.~Riquelme, J.~Puigcerver, B.~Mustafa, M.~Neumann, R.~Jenatton, A.~Susano~Pinto, D.~Keysers, and N.~Houlsby, ``Scaling vision with sparse mixture of experts,'' \emph{Advances in Neural Information Processing Systems}, vol.~34, pp. 8583--8595, 2021.

\bibitem{kahneman1973attention}
D.~Kahneman, \emph{Attention and effort}.\hskip 1em plus 0.5em minus 0.4em\relax Citeseer, 1973, vol. 1063.

\bibitem{hockey1997compensatory}
G.~R.~J. Hockey, ``Compensatory control in the regulation of human performance under stress and high workload: A cognitive-energetical framework,'' \emph{Biological psychology}, vol.~45, no. 1-3, pp. 73--93, 1997.

\bibitem{merlhiot2022drowsiness}
G.~Merlhiot and M.~Bueno, ``How drowsiness and distraction can interfere with take-over performance: A systematic and meta-analysis review,'' \emph{Accident Analysis \& Prevention}, vol. 170, p. 106536, 2022.

\bibitem{warm2008vigilance}
J.~S. Warm, R.~Parasuraman, and G.~Matthews, ``Vigilance requires hard mental work and is stressful,'' \emph{Human factors}, vol.~50, no.~3, pp. 433--441, 2008.

\bibitem{ma2018relationship}
J.~Ma, J.~Gu, H.~Jia, Z.~Yao, and R.~Chang, ``The relationship between drivers’ cognitive fatigue and speed variability during monotonous daytime driving,'' \emph{Frontiers in psychology}, vol.~9, p. 459, 2018.

\bibitem{paulhus2007self}
D.~L. Paulhus, S.~Vazire \emph{et~al.}, ``The self-report method,'' \emph{Handbook of research methods in personality psychology}, vol.~1, no. 2007, pp. 224--239, 2007.

\bibitem{zhang2018generalized}
Z.~Zhang and M.~Sabuncu, ``Generalized cross entropy loss for training deep neural networks with noisy labels,'' \emph{Advances in neural information processing systems}, vol.~31, 2018.

\bibitem{liu1994mental}
Y.~Liu and C.~D. Wickens, ``Mental workload and cognitive task automaticity: an evaluation of subjective and time estimation metrics,'' \emph{Ergonomics}, vol.~37, no.~11, pp. 1843--1854, 1994.

\bibitem{wang2025impact}
A.~Wang, W.~Shi, D.~He, and H.~Yang, ``Impact of multi-dimensional cognitive demands on takeover performance, physiological and eye-tracking measures in conditionally automated vehicles,'' \emph{Transportation Research Part F: Traffic Psychology and Behaviour}, vol. 114, pp. 461--475, 2025.

\bibitem{jaeggi2010concurrent}
S.~M. Jaeggi, M.~Buschkuehl, W.~J. Perrig, and B.~Meier, ``The concurrent validity of the n-back task as a working memory measure,'' \emph{Memory}, vol.~18, no.~4, pp. 394--412, 2010.

\bibitem{liang2010combining}
Y.~Liang and J.~D. Lee, ``Combining cognitive and visual distraction: Less than the sum of its parts,'' \emph{Accident Analysis \& Prevention}, vol.~42, no.~3, pp. 881--890, 2010.

\bibitem{tong2024difference}
Y.~Tong, B.~Jia, S.~Bao, C.~Wu, and N.~Sethuraman, ``The difference in physical and mental fatigue development between novice young adult and experienced middle-aged adult drivers during simulated automated driving,'' \emph{Journal of Safety Research}, vol.~91, pp. 165--174, 2024.

\bibitem{figalova2023fatigue}
N.~Figalov{\'a}, H.-J. Bieg, M.~Schulz, J.~Pichen, M.~Baumann, L.~L. Chuang, and O.~Pollatos, ``Fatigue and mental underload further pronounced in l3 conditionally automated driving: Results from an eeg experiment on a test track,'' in \emph{Companion Proceedings of the 28th International Conference on Intelligent User Interfaces}, 2023, pp. 64--67.

\bibitem{hart1988development}
S.~G. Hart and L.~E. Staveland, ``Development of nasa-tlx (task load index): Results of empirical and theoretical research,'' in \emph{Advances in psychology}.\hskip 1em plus 0.5em minus 0.4em\relax Elsevier, 1988, vol.~52, pp. 139--183.

\bibitem{aakerstedt1990subjective}
T.~{\AA}kerstedt and M.~Gillberg, ``Subjective and objective sleepiness in the active individual,'' \emph{International journal of neuroscience}, vol.~52, no. 1-2, pp. 29--37, 1990.

\bibitem{campbell2012morepower}
J.~I. Campbell and V.~A. Thompson, ``Morepower 6.0 for anova with relational confidence intervals and bayesian analysis,'' \emph{Behavior research methods}, vol.~44, pp. 1255--1265, 2012.

\bibitem{ahlstrom2024development}
C.~Ahlstr{\"o}m and A.~Anund, ``Development of sleepiness in professional truck drivers: Real-road testing for driver drowsiness and attention warning (ddaw) system evaluation,'' \emph{Journal of sleep research}, p. e14259, 2024.

\bibitem{yang2021impacts}
S.~Yang, J.~Kuo, M.~G. Lenn{\'e}, M.~Fitzharris, T.~Horberry, K.~Blay, D.~Wood, C.~Mulvihill, and C.~Truche, ``The impacts of temporal variation and individual differences in driver cognitive workload on ecg-based detection,'' \emph{Human factors}, vol.~63, no.~5, pp. 772--787, 2021.

\bibitem{revanur2021first}
A.~Revanur, Z.~Li, U.~A. Ciftci, L.~Yin, and L.~A. Jeni, ``The first vision for vitals (v4v) challenge for non-contact video-based physiological estimation,'' in \emph{Proceedings of the IEEE/CVF international conference on computer vision}, 2021, pp. 2760--2767.

\bibitem{wei2025prior}
F.~Wei, J.~Yang, Y.~Wang, L.~Lin, and H.~Zhang, ``Prior knowledge-guided multi-information graph convolutional network for driver drowsiness detection,'' \emph{Expert Systems with Applications}, vol. 275, p. 127028, 2025.

\bibitem{hara3dcnns}
K.~Hara, H.~Kataoka, and Y.~Satoh, ``Can spatiotemporal 3d cnns retrace the history of 2d cnns and imagenet?'' in \emph{Proceedings of the IEEE Conference on Computer Vision and Pattern Recognition (CVPR)}, 2018, pp. 6546--6555.

\bibitem{arnab2021vivit}
A.~Arnab, M.~Dehghani, G.~Heigold, C.~Sun, M.~Lu{\v{c}}i{\'c}, and C.~Schmid, ``Vivit: A video vision transformer,'' in \emph{Proceedings of the IEEE/CVF international conference on computer vision}, 2021, pp. 6836--6846.

\bibitem{niu2019rhythmnet}
X.~Niu, S.~Shan, H.~Han, and X.~Chen, ``Rhythmnet: End-to-end heart rate estimation from face via spatial-temporal representation,'' \emph{IEEE Transactions on Image Processing}, vol.~29, pp. 2409--2423, 2019.

\bibitem{wang2024multi}
J.~Wang, A.~Wang, H.~Hu, K.~Wu, and D.~He, ``Multi-source domain generalization for ecg-based cognitive load estimation: Adversarial invariant and plausible uncertainty learning,'' in \emph{ICASSP 2024-2024 IEEE International Conference on Acoustics, Speech and Signal Processing (ICASSP)}.\hskip 1em plus 0.5em minus 0.4em\relax IEEE, 2024, pp. 1631--1635.

\bibitem{wang2024genelizable}
J.~Wang, H.~Lu, H.~Han, Y.~Chen, D.~He, and k.~Wu, ``Generalizable remote physiological measurement via semantic-sheltered alignment and plausible style randomization,'' \emph{IEEE Transactions on Instrumentation and Measurement}, 2024.

\bibitem{kim2022study}
D.-Y. Kim, S.-Y. Cho, K.~Lee, and C.-B. Sohn, ``A study of projection-based attentive spatial--temporal map for remote photoplethysmography measurement,'' \emph{Bioengineering}, vol.~9, no.~11, p. 638, 2022.

\bibitem{othman2022contactless}
W.~Othman, A.~Kashevnik, I.~Ryabchikov, and N.~Shilov, ``Contactless camera-based approach for driver respiratory rate estimation in vehicle cabin,'' in \emph{Proceedings of SAI Intelligent Systems Conference}.\hskip 1em plus 0.5em minus 0.4em\relax Springer, 2022, pp. 429--442.

\end{thebibliography}

\begin{IEEEbiography}[{\includegraphics[width=1in,height=1.25in,clip,keepaspectratio]{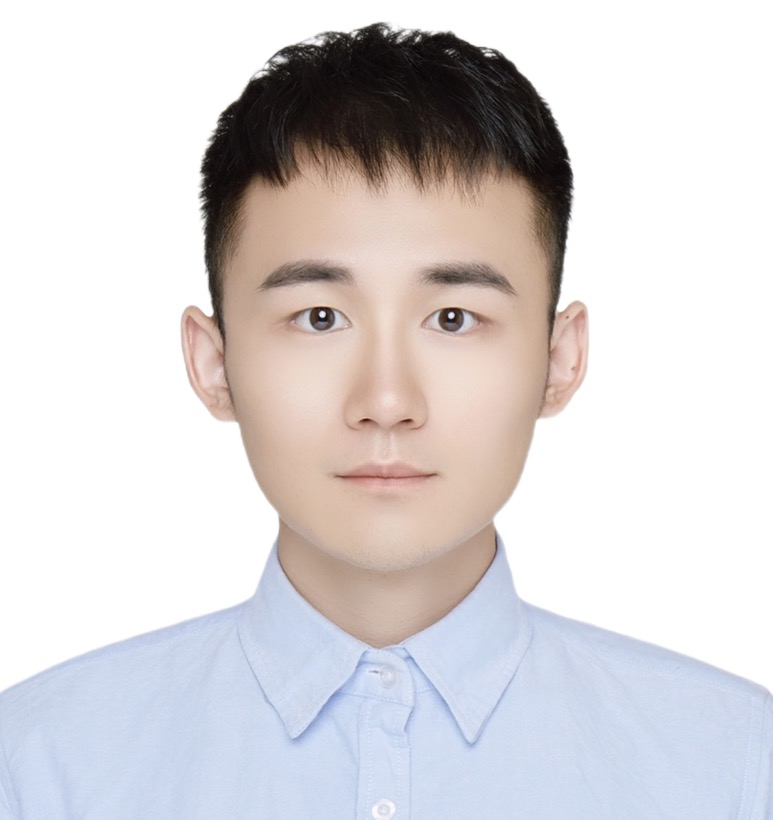}}]{Jiyao Wang}
received the B.Eng. degree in Software Engineering from Sichuan University, Chengdu, China in 2021, M.Sc. degree in Big Data Technology from the Hong Kong University of Science and Technology (HKUST), Hong Kong S.A.R., China, in 2022, and a Ph.D. degree at HKUST, Guangzhou campus. His research interests include physiological signal measurement, intelligent transport systems, and human factors.
\end{IEEEbiography}

\begin{IEEEbiography}[{\includegraphics[width=1in,height=1.25in,clip,keepaspectratio]{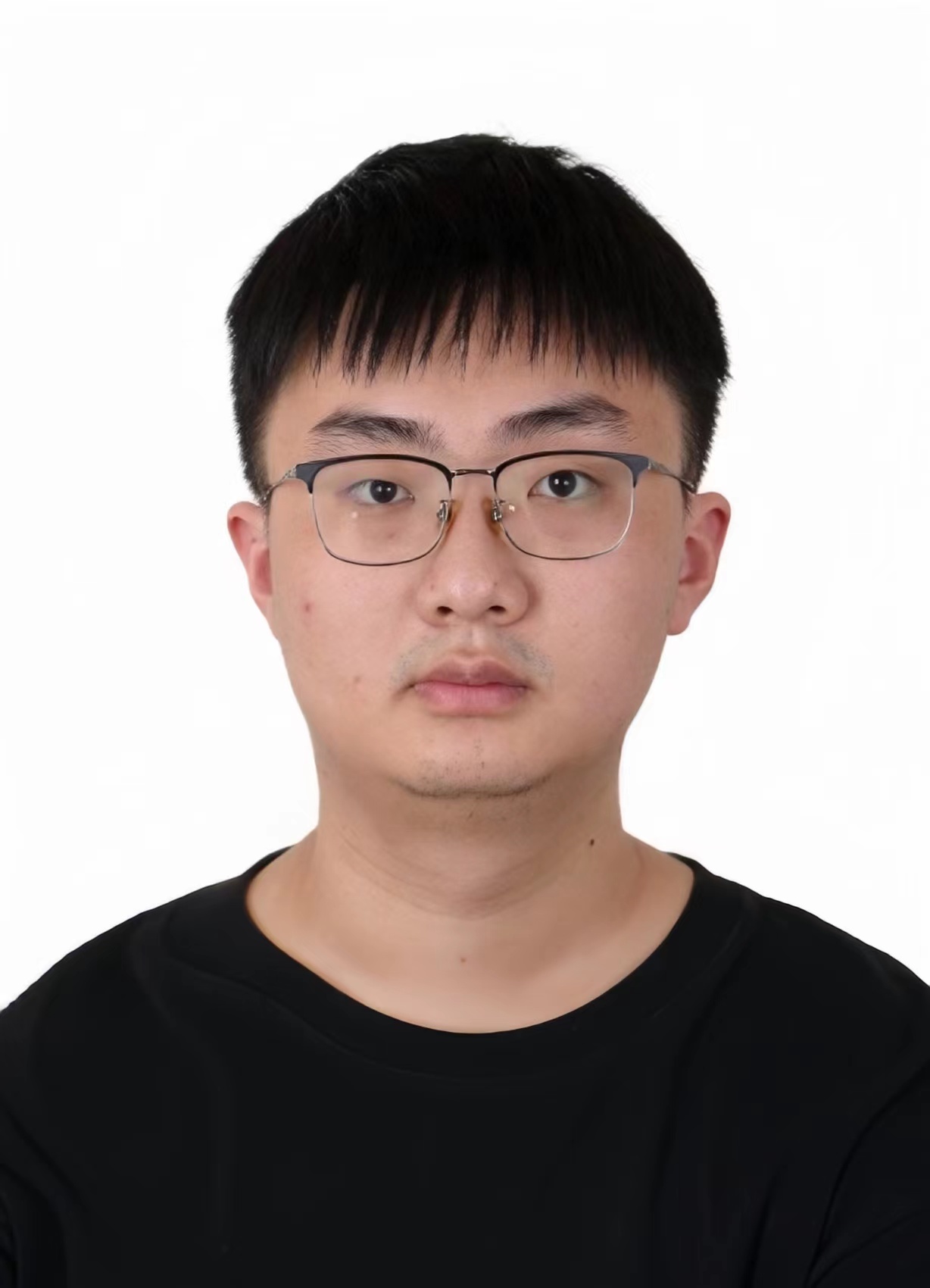}}]{Xiao Yang} obtained a bachelor's degree in Computing Science at Sichuan Agricultural University and is currently pursuing master degree at HKUST, Guangzhou campus. His research interests include physiological signal measurement, human factors, and remote sensing.
\end{IEEEbiography}

\begin{IEEEbiography}[{\includegraphics[width=1in,height=1.25in,clip,keepaspectratio]{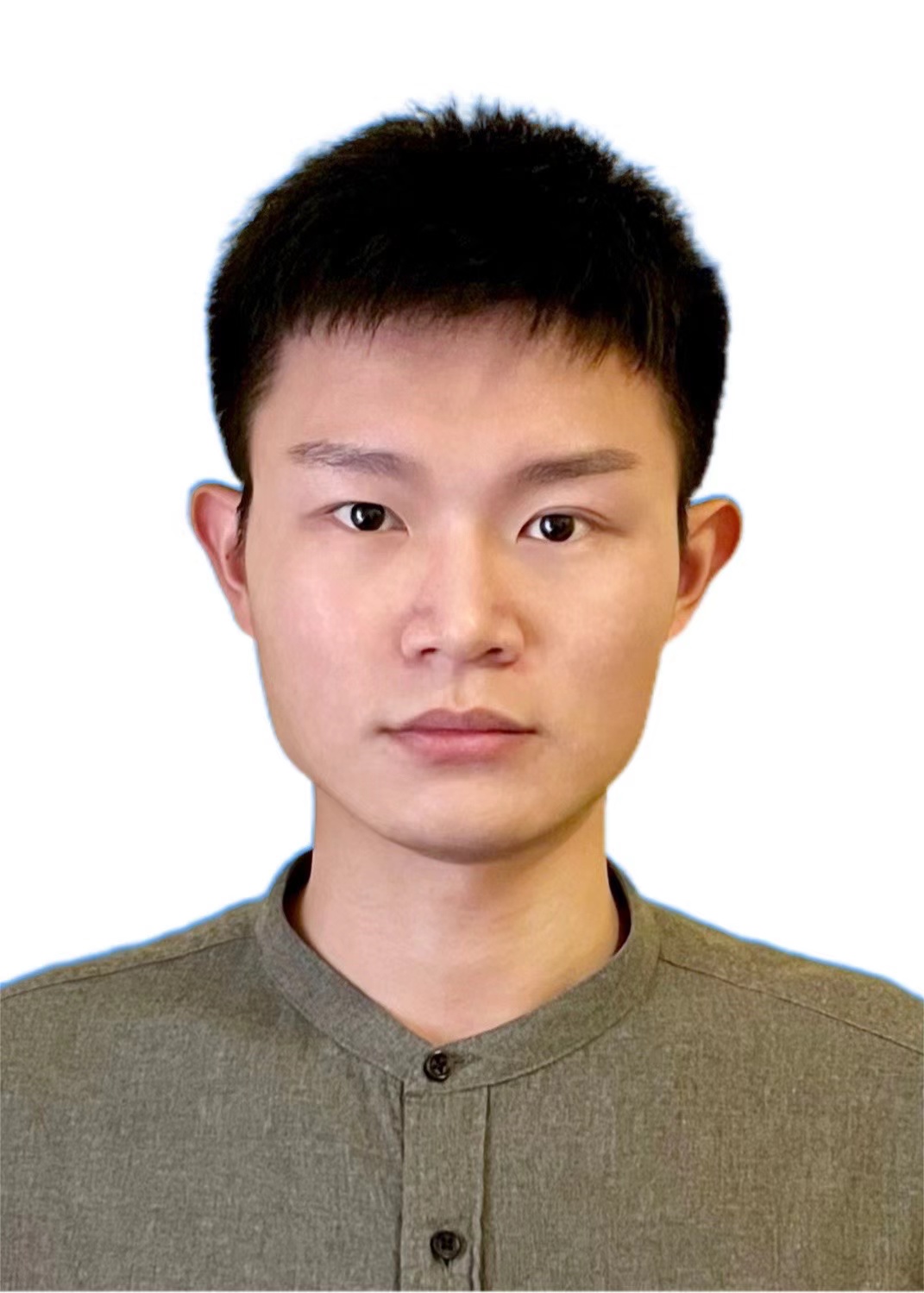}}]{Zhenyu Wang} received the B.S. degree in civil engineering from the University of Jinan and the M.S. degree in energy and power engineering from the Sun Yat-sen University. He is currently pursuing the Ph.D. degree with HKUST(GZ). His current research interests include human factors, thermal comfort, and driving safety.
\end{IEEEbiography}

\begin{IEEEbiography}[{\includegraphics[width=1in,height=1.25in,clip,keepaspectratio]{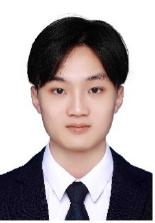}}]{Ximeng Wei} received his bachelor's degree in Refrigeration and Cryogenic engineering from Central South University, China in 2022. For now, he is pursuing the master's degree at the University of Hong Kong, and is the founder of the Hunch Innovation Innovation Technology Co., Ltd, Shenzhen, China. 
\end{IEEEbiography}

\begin{IEEEbiography}[{\includegraphics[width=1in,height=1.25in,clip,keepaspectratio]{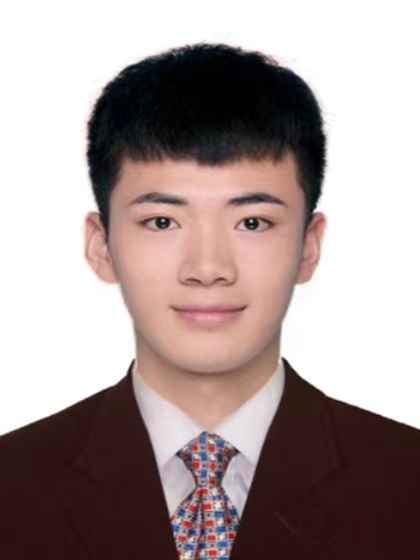}}]{Ange Wang}
received his bachelor's degree in traffic engineering from East China Jiaotong University in 2019 and M.Sc. degree in communication and transportation engineering from the Beijing University of Technology in 2022. He is now a Ph.D. student in the Intelligent Transportation Thrust at the Hong Kong University of Science and Technology (Guangzhou).
\end{IEEEbiography}

\begin{IEEEbiography}[{\includegraphics[width=1in,height=1.25in,clip,keepaspectratio]{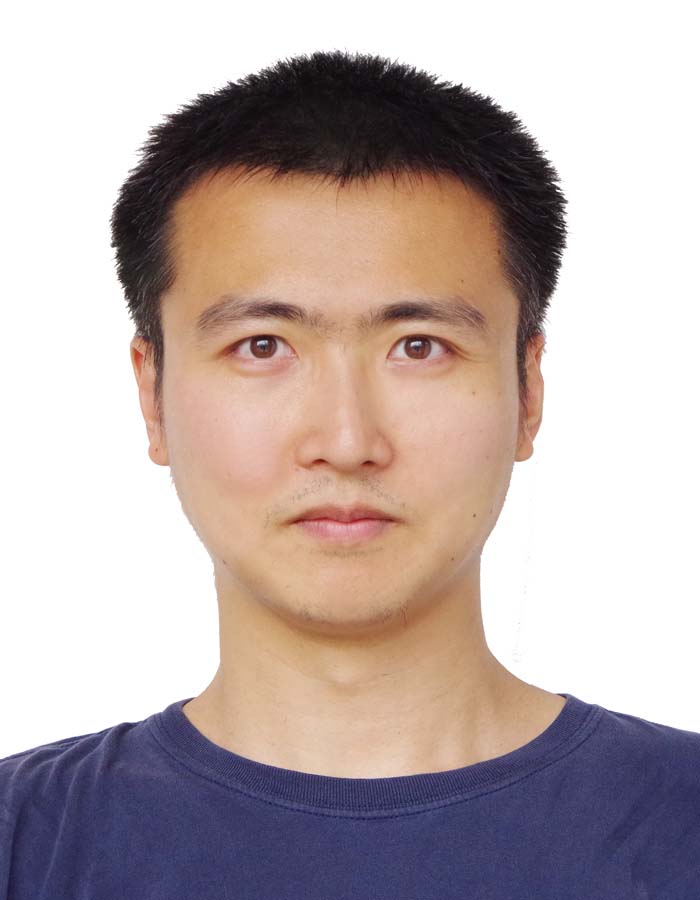}}]{Dengbo He} received his bachelor’s degree from Hunan University in 2012, M.S. degree from the Shanghai Jiao Tong University in 2016 and Ph.D. degree from the University of Toronto in 2020. He is currently an assistant professor from the Intelligent Transpiration Trust and Robotics and Autonomous Systems Thrust, the HKUST(Guangzhou). He is also affiliated with the Department of Civil and Environmental Engineering, HKUST, Hong Kong SAR. From 2020 to 2021, he was a post-doctoral fellow at the University of Toronto.
\end{IEEEbiography}

\begin{IEEEbiography}[{\includegraphics[width=1in,height=1.25in,clip,keepaspectratio]{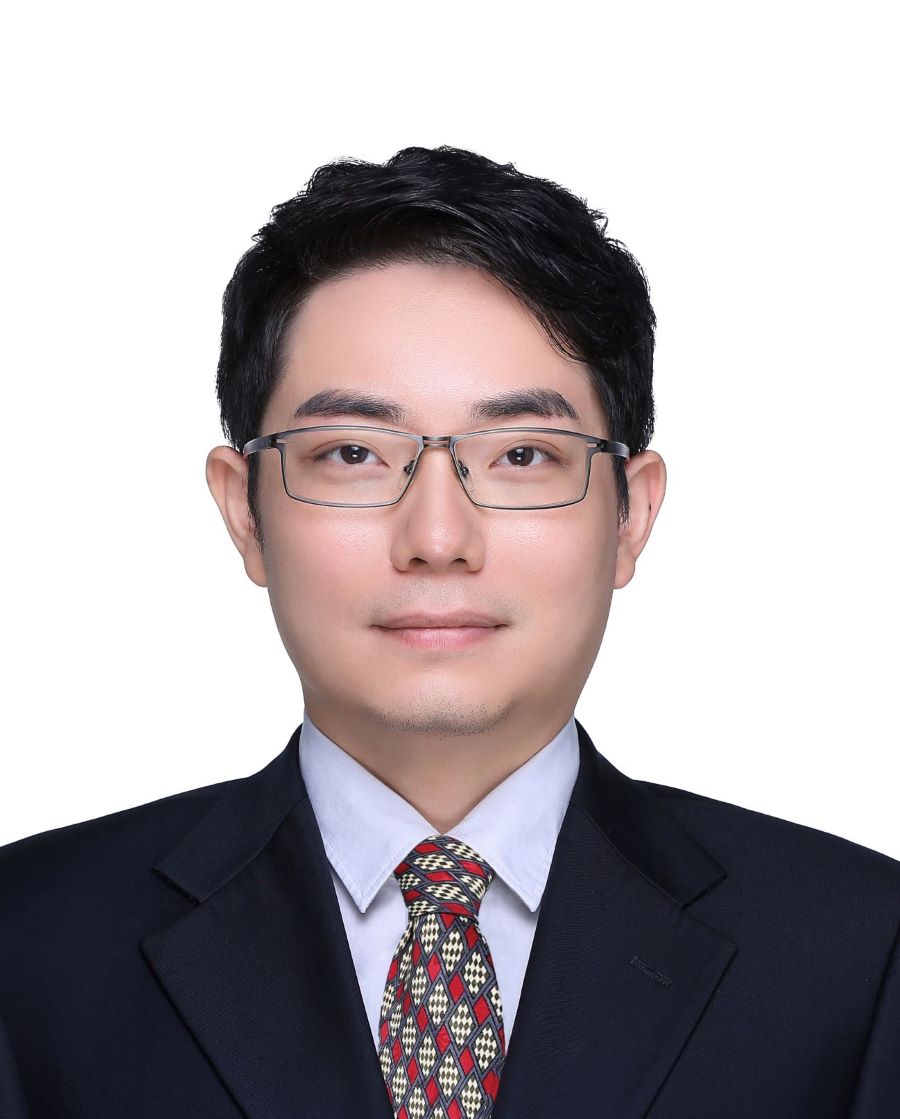}}]{Kaishun Wu}
(Fellow, IEEE) received the Ph.D. degree in computer science and engineering from HKUST, Hong Kong, in 2011. He was a Distinguished Professor and the Director of Guangdong Provincial Wireless Big Data and Future Network Engineering Center with Shenzhen University, Shenzhen, China. In 2022, he joined HKUST (GZ) as a Full Professor with DSA Thrust and IoT Thrust. He is an Active Researcher with more than 200 papers published on major international academic journals and conferences, as well as more than 100 invention patents, including 12 from the USA. He is an IET, AAIA, and IEEE Fellow.
\end{IEEEbiography}

\end{document}